# Data-Efficient Off-Policy Policy Evaluation for Reinforcement Learning

**Philip S. Thomas**  PHILIPT@CS.CMU.EDU
Carnegie Mellon University

**Emma Brunskill**  EBRUN@CS.CMU.EDU
Carnegie Mellon University

## Abstract

In this paper we present a new way of predicting the performance of a reinforcement learning policy given historical data that may have been generated by a different policy. The ability to evaluate a policy from historical data is important for applications where the deployment of a bad policy can be dangerous or costly. We show empirically that our algorithm produces estimates that often have orders of magnitude lower mean squared error than existing methods—it makes more efficient use of the available data. Our new estimator is based on two advances: an extension of the doubly robust estimator (Jiang & Li, 2015), and a new way to mix between model based estimates and importance sampling based estimates.

## 1. Introduction

The ability to predict the performance of a policy without actually having to use it is crucial to the responsible use of reinforcement learning algorithms. Consider the setting where the user of a reinforcement learning algorithm has already deployed some policy, e.g., for determining which advertisement to show a user visiting a website (Theocharous et al., 2015), for determining which medical treatment to suggest for a patient (Thapa et al., 2005), or for suggesting a personalized curriculum for a student (Mandel et al., 2014). In these examples, using a bad policy can be costly or dangerous, so it is important that the user of a reinforcement learning algorithm be able to accurately predict how well a new policy will perform without having to deploy it.

In this paper we propose a new algorithm for tackling this performance prediction problem, which is called the *off-policy policy evaluation* (OPE) problem. The primary objective in OPE problems is to produce estimates that minimize some notion of error. We select mean squared error, a popular notion of error for estimators, as our loss function. This is in line with previous works that all use (root) mean squared error when empirically validating their methods (Precup et al., 2000; Dudík et al., 2011; Mahmood et al., 2014; Thomas, 2015b; Jiang & Li, 2015).

Given this goal, an estimator should be strongly consistent—its mean squared error should converge almost surely to zero as the amount of available data increases.[1] In this paper we introduce a new strongly consistent estimator, MAGIC, that directly optimizes mean squared error. Our empirical results show that MAGIC can produce estimates with orders of magnitude lower mean squared error than the estimates produced by existing algorithms.

Our new algorithm comes from the synthesis of two new contributions. The first contribution is an extension of the recently proposed *doubly robust* (DR) OPE algorithm (Jiang & Li, 2015). We present a novel derivation of their algorithm that removes the assumption that the horizon is finite and known. We also give conditions under which the DR estimator is strongly consistent. We then show how we can significantly reduce the variance of the DR estimator by introducing a small amount of bias—an effective trade-off when attempting to minimize the mean squared error of the estimates. We call our extension of the DR estimator the *weighted doubly robust* (WDR) estimator.

Our second major contribution is a new estimator, which we call the *blending IS and model* (BIM) estimator, that combines two different OPE estimators not just by selecting between them, but by blending them together in a way that minimizes the mean squared error. The combination of these two contributions results in a particularly powerful new OPE algorithm that we call the *model and guided importance sampling combining* (MAGIC) estimator, which uses BIM to combine a purely model-based estimator with WDR. In our simulations, MAGIC has the best general per-formance, often exhibiting orders of magnitude lower mean squared error than prior state-of-the-art estimators.

---

[1] A formal definition of what it means for an estimator to be strongly consistent is provided in Appendix A, where, in Lemma 3, we also elucidate the relationship between strong consistency and mean squared error.

## 2. Notation

We assume that the reader is familiar with reinforcement learning (Sutton & Barto, 1998) and adopt notational standard MDPNv1 for *Markov decision processes* (Thomas, 2015a, MDPs). For simplicity, our notation assumes that the state, action, and reward sets are finite, although our results carry over to more general settings.[2] Let $H := (S_0, A_0, R_0, S_1, \dots)$ be a *trajectory*,[3] and $g(H) := \sum_{t=0}^{\infty} \gamma^t R_t$ denote the *return* of a trajectory. We assume that the (possibly unknown) minimum and maximum rewards, $r_{\min}$ and $r_{\max}$, are finite and that $\gamma \in [0, 1]$ for the finite-horizon setting and $\gamma \in [0, 1)$ for the indefinite and infinite horizon settings so that $g(H)$ is bounded. We use the discounted objective function, $v(\pi) := \mathbf{E}[g(H)|H \sim \pi]$, where $H \sim \pi$ denotes that $H$ was generated using the policy $\pi$. When dealing with multiple trajectories, we use superscripts to denote which trajectory a term comes from. For example, we write $S_t^H$ to denote the state at time $t$ during trajectory $H$. Let $v^\pi$ and $q^\pi$ be the *state value function* and *state-action value function* for policy $\pi$—for all $(\pi, s, a) \in \Pi \times \mathcal{S} \times \mathcal{A}$, let $v^\pi(s) := \mathbf{E}\left[\sum_{t=0}^{\infty} \gamma^t R_t | S_0 = s, \pi\right]$ and $q^\pi(s, a) := \mathbf{E}\left[\sum_{t=0}^{\infty} \gamma^t R_t | S_0 = s, A_0 = a, \pi\right]$. Notice that $v$ without a superscript denotes the objective function, while $v^\pi$ denotes a value function.

We will assume that *historical data*, $D$, is provided. Formally this historical data is a set of $n \in \mathbb{N}_{>0}$ trajectories and the known policies, called *behavior policies*, that were used to generate them. That is, $D := \{(H_i, \pi_i)\}_{i=1}^n$, where $H_i \sim \pi_i$. Importantly, when we write $H_i$, we *always* mean that $H_i \sim \pi_i$. Let $\rho_t(H, \pi_e, \pi_b) := \prod_{i=0}^{t} \pi_e\left(A_i^H | S_i^H\right) / \pi_b\left(A_i^H | S_i^H\right)$, be an *importance weight*, which is the probability of the first $t$ steps of $H$ under the *evaluation policy*, $\pi_e$, divided by its probability under the behavior policy $\pi_b$ (Precup et al., 2000, Section 2). For brevity, we write $\rho_t^i$ as shorthand for $\rho_t(H_i, \pi_e, \pi_i)$ and $\rho_t$ as shorthand for $\rho_t(H, \pi_e, \pi_b)$. To simplify later expressions, let $\rho_{-1}^i := 1$ for all $i$. One of the primary challenges will be to combat the high variance and large range of the importance weights, $\rho_t$.

Some of the methods that we describe use an approximate model of an MDP. Let $\hat{r}^\pi(s, a, t)$ denote the model's prediction of the reward $t$ steps later, $R_t$, if $S_0 = s$, $A_0 = a$, and the policy $\pi$ is used to generate the subsequent actions, $A_1, A_2, \dots$. For example, $\hat{r}^\pi(s, a, 0)$ is a prediction of the immediate reward after taking action $a$ in state $s$ and is thus the same for all policies, $\pi$. We assume that these predictions are bounded by finite (possibly unknown) constants $r_{\min}^{\text{model}}$ and $r_{\max}^{\text{model}}$, i.e., $\hat{r}^\pi(s, a, t) \in [r_{\min}^{\text{model}}, r_{\max}^{\text{model}}]$. Let

$$\hat{r}^\pi(s, t) := \sum_{a \in \mathcal{A}} \pi(a|s)\hat{r}^\pi(s, a, t), \qquad (1)$$

be a prediction of $R_t$ if $S_0 = s$ and the policy $\pi$ is used to generate actions $A_0, A_1, \dots$, for all $(s, t, \pi) \in \mathcal{S} \times \mathbb{N}_{\geq 0} \times \Pi$. Let $\hat{v}^\pi(s) := \sum_{t=0}^{\infty} \gamma^t \hat{r}^\pi(s, t)$ and $\hat{q}^\pi(s, a) := \sum_{t=0}^{\infty} \gamma^t \hat{r}^\pi(s, a, t)$ be the model's estimates of $v^\pi(s)$ and $q^\pi(s, a)$. We assume that if a *terminal absorbing state*, $\overset{\infty}{s}$, is reached, the model's predictions of rewards that occur thereafter are always zero: $\hat{r}^\pi(\overset{\infty}{s}, a, t) = 0$ for all $(\pi, a, t) \in \Pi \times \mathcal{A} \times \mathbb{N}_{\geq 0}$. Although better models will tend to improve our estimates, we make no assumptions about the veracity of the approximate model's predictions, $\hat{r}^\pi(s, a, t)$.

## 3. Off-Policy Policy Evaluation (OPE)

The problem of *off-policy policy evaluation* (OPE) is defined as follows. We are given an evaluation policy, $\pi_e$, *historical data*, $D$, and an approximate model. Our goal is to produce an estimator, $\hat{v}(D)$, of $v(\pi_e)$ that has low *mean squared error* (MSE): $\text{MSE}(\hat{v}(D), v(\pi_e)) := \mathbf{E}\left[(\hat{v}(D) - v(\pi_e))^2\right]$. We use capital letters to denote random variables, and so the random terms in expected values are always the capitalized letters (e.g. $D$ is a random variable). We assume that the process producing states, actions, and rewards is an MDP with an unknown initial state distribution, transition function, and reward function. We assume that the evaluation policy, $\pi_e$, the behavior policies, $\pi_i, i \in \{1, \dots, n\}$, and the discount parameter, $\gamma$, are known. For a review of OPE methods, see the works of Precup et al. (2000) or Thomas (2015b, Chapter 3). More recent methods can be found in the works of Jiang & Li (2015) and Mandel et al. (2016).

## 4. Doubly Robust (DR) Estimator

The *doubly robust* (DR) estimator (Jiang & Li, 2015) is a new unbiased estimator of $v(\pi_e)$ that achieves promising empirical and theoretical results by leveraging an approximate model of an MDP to decrease the variance of the unbiased estimates produced by ordinary importance sam-pling (Precup et al., 2000). It is doubly robust in that it will provide "good" estimates if either **1)** the model is accurate or **2)** the behavior policies are known. By "good" it is meant that if the former does not hold then the estimator will remain unbiased (although it might have high variance and thus high mean squared error), and if the latter does not hold then if the model has low error the doubly robust estimator will also tend to have low error. Doubly robust estimators were introduced and remain popular in the statistics community (Rotnitzky & Robins, 1995; Heejung & Robins, 2005).

---

[2] We have verified that our results carry over to the setting where the states, actions, and rewards are continuous random variables with density functions. This result is relatively straightforward—summations are replaced with integrals, probability mass functions with probability density functions, and probabilities with probability densities, where appropriate. We have included Assumption 4, which is implied when the setting is restricted to finite state, action, and reward sets, since it is necessary for the continuous setting. We have *not* verified our results for the setting where the states, actions, and rewards come from distributions that do not have probability mass or density functions, e.g., if the rewards come from the Cantor distribution.

[3] For alliteration, one might think of $H$ as denoting the *history of an episode*.

The work that introduced the DR estimator for MDPs (Jiang & Li, 2015) derived it as a generalization of a doubly robust estimator for bandits (Dudík et al., 2011). This may be why the DR estimator was derived only for the finite horizon setting where the horizon is known (every trajectory must terminate within $L < \infty$ time steps, and $L$ must be known). It also resulted in a recursive definition of the DR estimator that can be difficult to interpret. In Appendix B we instead derive the DR estimator for MDPs as an application of control variates. Our new derivation holds without assumptions on the horizon and gives the intuitive non-recursive definition, where $w_t^i = \rho_t^i/n$:

$$\mathrm{DR}(D) := \sum_{i=1}^n \sum_{t=0}^\infty \gamma^t w_t^i R_t^{H_i} \qquad (2)$$
$$- \sum_{i=1}^n \sum_{t=0}^\infty \gamma^t \left( w_t^i \hat{q}^{\pi_e}\left(S_t^{H_i}, A_t^{H_i}\right) - w_{t-1}^i \hat{v}^{\pi_e}\left(S_t^{H_i}\right) \right).$$

In Appendix B we show that this definition is equivalent to that of Jiang & Li (2015) when the horizon is finite and known, and we provide several new theoretical results pertaining to the DR estimator. Specifically, we give conditions for DR to be an unbiased estimator without assumptions on the horizon, and we give the first proofs that it is a strongly consistent estimator. Although these are important properties to establish, we relegate them to an appendix due to space limitations.

The non-recursive definition of the DR estimator presented in (2) also reveals the close relationship of the DR estimator to *advantage sum* estimators. Advantage sum estimators were introduced as a way to lower the variance of on-policy Monte Carlo performance estimates for a setting that is a generalization of the (partially observable) MDP setting (Zinkevich et al., 2006; White & Bowling, 2009). The DR estimator for the on-policy setting can be found in the work of Zinkevich et al. (2006, Equation 8). One may therefore view the DR estimator (Jiang & Li, 2015) as the extension of the advantage sum estimator (Zinkevich et al., 2006) to the off-policy setting or as the extension of the doubly robust estimator for bandits (Dudík et al., 2011) to the sequential setting. We are therefore not the first to show that the DR estimator can be viewed as an application of control variates, since Veness et al. (2011, Section 3.1) point out that the advantage sum estimator is an application of control variates. Still, our derivation in Appendix B of the DR estimator is novel.

The DR estimator is not purely model based, since it uses importance weights. However, it is also not a model-free importance sampling method, since it uses an approximate model to decrease the variance of its estimates. We therefore refer to it as a *guided importance sampling* method, since the approximate model is used to guide, but not completely replace, the importance sampling estimates.

## 5. Weighted Doubly Robust (WDR) Estimator

Empirical and theoretical results show that the DR estimator developed, analyzed, and tested by Jiang & Li (2015) can significantly reduce the variance of ordinary importance sampling without introducing bias. The fact that it does not introduce bias can be particularly important when the estimator is used to produce confidence bounds on $v(\pi_e)$ (Thomas, 2015b). However, in practice these confidence bounds often require an impractical amount of data before they are tight enough to be useful, and so approximate confidence bounds (e.g., bootstrap confidence bounds) are used instead (Theocharous et al., 2015; Thomas, 2015b). When using these approximate confidence bounds, the strict requirement that an OPE estimator be an unbiased estimator of $v(\pi_e)$ is not necessary. Furthermore, often the goal of OPE is not to produce confidence bounds, but to produce the best estimate of $v(\pi_e)$ possible, in order to determine whether $\pi_e$ should be used instead of the current behavior policy or as an internal mechanism in a policy search algorithm (Levine & Koltun, 2013). In these cases, the "best" estimator is typically defined as the one that has the lowest *mean squared error* (MSE). For example, in their experiments, Precup et al. (2000), Dudík et al. (2011), Mahmood et al. (2014), Thomas (2015b), and Jiang & Li (2015) all use the (root) MSE when evaluating OPE methods.

Although unbiasedness might seem like a desirable property of an estimator, when the goal is to minimize MSE, it often is not. In general, the MSE of an estimator, $\hat{\theta}$, of a statistic, $\theta$, can be decomposed into its variance and its squared bias: $\mathrm{MSE}(\hat{\theta}, \theta) = \mathbf{E}[(\theta - \hat{\theta})^2] = \mathrm{Var}(\hat{\theta}) + \mathrm{Bias}(\hat{\theta})^2$, where $\mathrm{Bias}(\hat{\theta}) := \mathbf{E}[\hat{\theta}] - \theta$. The optimal estimator in terms of MSE is typically one that balances this bias-variance trade-off, not one with zero bias. Therefore, in the context of minimizing MSE, strong asymptotic consistency, which requires the MSE of an estimator to almost surely converge to zero as the amount of available data increases, is a more desirable property than unbiasedness.

In this section we propose a new OPE estimator that we call the *weighted doubly robust* (WDR) estimator. The WDR estimator comes from applying a simple well-known extension to importance sampling estimators to the DR estimator to produce a new guided importance sampling method. This extension does not directly optimize the bias-variance trade-off, but it does tend to significantly better balance it while maintaining asymptotic consistency. More specifically, WDR is based on *weighted importance sampling* (Powell & Swann, 1966) as opposed to ordinary importance sampling (Hammersley & Handscomb, 1964). For further discussion of the benefits of weighted importance sampling over ordinary importance sampling, see the work of Thomas (2015b, Section 3.8). Weighted importance sampling has been used before for OPE (Precup et al., 2000), but not in conjunction with the DR estimator.

Our WDR estimator is defined as the DR estimator in (2), except where $w_t^i := \rho_t^i / \sum_{j=1}^n \rho_t^j$.[4] Intuitively it is clear that this estimator is asymptotically correct because $\mathbf{E}[\rho_t^j] = 1$, and so by the law of large numbers the denominator of $w_t^i$ will converge to $n$. Although WDR is not an unbiased estimator of $v(\pi_e)$, its bias follows a pattern that is both predictable and also sometimes desirable. When there is only a single trajectory, i.e., $n = 1$, WDR($D$) is an unbiased estimator of the performance of the behavior policy, since $w_t^1 = 1$ for all $t$. If there is a single behavior policy, $\pi_b$, as the number of trajectories increases, the expected value of WDR($D$) shifts away from $v(\pi_b)$ towards $v(\pi_e)$.

Before presenting theoretical results about the WDR estimator, we introduce assumptions that they will require. These assumptions are only included if they are explicitly mentioned in a theorem—most theorems only rely on few of these assumptions. Even when these assumptions are not satisfied, it does *not* mean that the result does not hold or that the WDR estimator will perform poorly—it merely means that the theoretical results that we provide are not guaranteed by our proofs.

Assumption 1 ensures that all trajectories of interest when evaluating $\pi_e$ will be produced by all of the behavior policies. This is a standard assumption in OPE and typically precludes the use of deterministic behavior policies.[5]

**Assumption 1** (Absolute continuity). *For all* $(s, a, i) \in$ $\mathcal{S} \times \mathcal{A} \times \{1, \ldots, n\}$, *if* $\pi_i(a|s) = 0$ *then* $\pi_e(a|s) = 0$.

Assumption 2 requires all of the behavior policies to be identical. This assumption is trivially satisfied if data is collected from one behavior policy. Also, this assumption is often satisfied for applications where data is abundant, so that evaluation can be performed using only the data from the most recent behavior policy. Also, Assumption 3 requires the horizon, $L$, to be finite.

**Assumption 2** (Single behavior policy). *For all* $(i, j) \in \{1, \ldots, n\}^2$, $\pi_i = \pi_j$.

**Assumption 3.** *$L$ is finite.*

Assumption 4 requires the importance weights, $\rho_t^i$, to be bounded above by a finite constant, $\beta \in \mathbb{R}$ (they are always bounded below by zero). It is trivially satisfied in the common setting where the horizon is finite and the state and action sets are finite. Although Assumption 4 requires $\beta$ to exist, none of our results depend on how large $\beta$ is. So, in the non-finite state, action, and horizon settings one may ensure that evaluation policies are only considered if they satisfy Assumption 4 for some arbitrarily large $\beta$.

**Assumption 4** (Bounded importance weight). *There exists a constant $\beta < \infty$ such that for all $(t, i) \in \mathbb{N}_{\geq 0} \times \{1, \ldots, n\}$, $\rho_t^i \leq \beta$ surely.*

In the following theorems, Theorems 1 and 2, we give two different sets of assumptions that are sufficient to show that WDR is a strongly consistent estimator of $v(\pi_e)$. Notice that if the sets of states and actions are finite and the horizon is finite, then Assumption 4 holds, and so Theorem 2 means that WDR will be strongly consistent given only Assumption 1.

**Theorem 1** (WDR – strongly consistent estimator for one behavior policy, finite horizon). *If Assumptions 1, 2, and 3 hold then* WDR($D$) $\xrightarrow{a.s.}$ $v(\pi_e)$. **Proof** *See Appendix C.1.*

**Theorem 2** (WDR – strongly consistent estimator for many behavior policies). *If Assumptions 1 and 4 hold then* WDR($D$) $\xrightarrow{a.s.}$ $v(\pi_e)$. **Proof** *See Appendix C.2.*

## 6. Empirical Studies (WDR)

In order to both show the empirical benefits of WDR over existing importance sampling estimators and better motivate our second major contribution, in this section we present an empirical comparison of different OPE methods.[6] We compare to a broad sampling of model-free importance sampling estimators, definitions of which can be found in the work of Thomas (2015b, Chapter 3): *importance sampling* (IS), *per-decision importance sampling*

---

[4] Just as DR-v2 extends the DR estimator (Jiang & Li, 2015, Section 4.4), one can create the WDR-v2 estimator by replacing $\hat{q}^{\pi_e}(S_t, A_t)$ with $\hat{r}^{\pi_e}(S_t, A_t, 0) + \gamma \hat{v}^{\pi_e}(S_{t+1})$ in (2). Empirical results for DR-v2 and WDR-v2 are included in the spreadsheet in the supplemental documents. For the domains presented here, these variants did not outperform the original DR and WDR estimators.

[5] Assumption 1 could be replaced with a less-restrictive assumption like that used by Thomas (2015b, Section 3.5). We use Assumption 1 because it allows for simplified proofs.

[6] The raw data for all of the plots provided in this paper (as well as additional plots) are available in the spreadsheet included in the supplemental material.

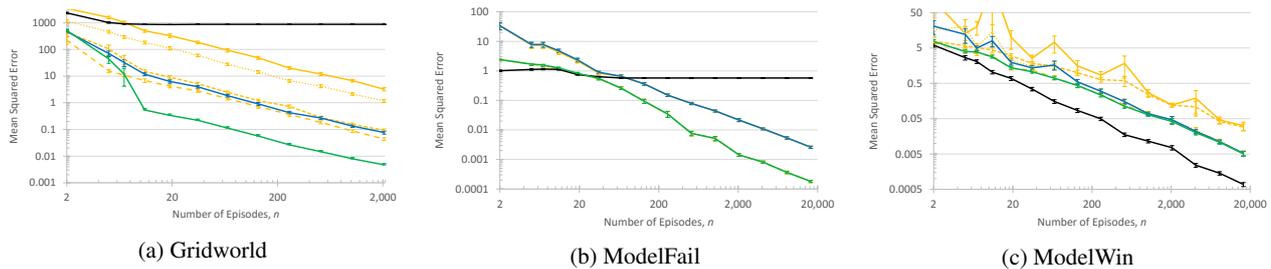

(a) Gridworld (b) ModelFail (c) ModelWin

Figure 1: Empirical results for three different experimental setups. All plots in this paper have the same format: they show the mean squared error of different estimators as $n$, the number of episodes in $D$, increases. Both axes always use a logarithmic scale and standard error bars are included from 128 trials. All plots use the following legend:

—— IS ········ PDIS - - - - WIS - - - CWPDIS —— DR —— AM —— WDR

(PDIS), *weighted importance sampling* (WIS), and *consistent weighted per-decision importance sampling* (CWPDIS). We also compare to the guided importance sampling *doubly robust* (DR) estimator (Jiang & Li, 2015).

Lastly, we compare to the *approximate model* (AM) estimator, which uses all of the available data to construct an approximate model of the MDP.[7] The performance of the evaluation policy on the approximate model is typically easy to compute and can be used as an estimate of $v(\pi_e)$. For example, in our experiments the approximate model maintains an estimate, $\widehat{d}_0$, of the initial state distribution, and so we define $\text{AM} \coloneqq \sum_{s \in \mathcal{S}} \widehat{d}_0(s) \hat{v}^{\pi_e}(s)$. Notice that unlike the importance sampling based methods, AM does not include any importance weights ($\rho_t$ terms).

In Appendix D we provide detailed descriptions of the experimental setup and results. Here we provide only an overview. We used three domains: a $4 \times 4$ gridworld previously constructed specifically for evaluating OPE methods (Thomas, 2015b, Section 2.5), as well as two simple domains that we developed to exemplify the settings where different methods excel and fail. In our simulations, WDR dominated the other importance sampling and guided importance sampling estimators (but not AM). Not only did WDR always achieve the lowest mean squared error of these estimators, but no other single (guided) importance sampling estimator was able to always achieve mean squared errors within an order of WDR's. Figure 1a is an example using the gridworld where no other method achieved mean squared errors within an order of WDR's.

The second notable trend is that WDR often significantly outperformed AM. We constructed a simple MDP and experimental setup that we call *ModelFail*, which exemplifies this. In the ModelFail experiments, the approximate model uses function approximation, which causes it to converge to the wrong MDP. This results in AM's mean squared error plateauing at a non-zero value, as shown in Figure 1b. Since WDR remains strongly consistent in this setting, its MSE converges almost surely to zero.

The third notable trend is that when an accurate approximate model is available, WDR does not always outperform AM. We constructed a simple MDP that we call *ModelWin*, which exemplifies this. In the ModelWin MDP, the approximate model quickly converges to the correct MDP, and so AM outperforms WDR. This is depicted in Figure 1c.

One might wonder why DR and WDR can do worse than AM even though they incorporate the approximate model. Notice that we can write the DR and WDR estimators as:

$$\text{WDR}(D) \coloneqq \underbrace{\frac{1}{n} \sum_{i=1}^{n} \hat{v}^{\pi_e}(S_0^{H_i})}_{(a)} \quad (3)$$
$$+ \sum_{i=1}^{n} \sum_{t=0}^{\infty} \gamma^t w_t^i \underbrace{\left[ R_t^{H_i} - \hat{q}^{\pi_e}\left(S_t^{H_i}, A_t^{H_i}\right) + \gamma \hat{v}^{\pi_e}\left(S_{t+1}^{H_i}\right) \right]}_{(b)}.$$

If the approximate model is perfect, then **(a)** is both a low variance and unbiased estimator of $v(\pi_e)$. If the approximate model is perfect and $R_t$ and $S_{t+1}$ are deterministic functions of $S_t$ and $A_t$, then **(b)** is zero, and so the second term is always zero and WDR is an excellent estimator. However, if $R_t$ or $S_{t+1}$ is *not* a deterministic function of $S_t$ and $A_t$—if the state transitions or rewards are stochastic—then **(b)** is not necessarily zero. If the importance weights, $w_t^i$, have high variance, then even slightly non-zero values of **(b)** can cause DR and WDR to have high variance.

In summary, in our experiments, WDR dominated the other (guided) importance sampling estimators, sometimes achieving orders of magnitude lower MSE. However, the experiments also show that WDR is not always the best

---
[7]This model-based estimator has been called the *direct method* in previous work (Dudík et al., 2011), however, in other previous work *direct methods* are model-free while *indirect methods* are model-based (Sutton & Barto, 1998, Section 9.2).

estimator—sometimes AM can produce estimates with an order lower MSE. This trend is also visible in the results of Jiang & Li (2015), where AM performs better than DR (although they did not compare to WDR, since it had not yet been introduced). Ideally we would like an estimator that combines WDR and AM or switches between them automatically, to always achieve the performance of the better estimator. In the following sections we show how this can be done.

## 7. Blending IS and Model (BIM) Estimator

In this section we show how two OPE estimators can be merged into a single estimator that exhibits the desirable properties of both. Before doing so, we establish some terminology. We divide OPE estimators into three classes. The first class we call *importance sampling estimators*. We define this class to include all estimators that, when $L$ is finite, are defined using all of the importance weights $\rho_0, \rho_1, \ldots, \rho_{L-1}$. Notice that this includes IS, PDIS, WIS, and CWPDIS, as well as the guided importance sampling methods, DR and WDR.

The second class we call *purely model-based estimators*. We define this class to include all estimators that do not contain any $\rho_t$ terms for $t \geq 0$. The only purely model-based estimator in this paper is AM. Finally, we call the third class *partial importance sampling estimators*. These estimators are those that do not fall into either of the other two classes—estimators that use importance weights, $\rho_t$, but only for $t < L - 1$. We will introduce one such estimator later in this section.

We contend that importance sampling estimators and purely model-based estimators are two extremes on a spectrum of estimators. Importance sampling estimators tend to be strongly consistent. That is, as more historical data becomes available, their estimates become increasingly accurate. However, their use of importance weights means that they *all* (including DR and WDR) also can have high variance relative to purely model-based estimators. This is evident in the results on the ModelWin domain.

On the other end of the spectrum, purely model-based estimators like AM are often *not* strongly consistent. If the approximate model uses function approximation or if there is some partial observability, then the approximate model may not converge to the true MDP. So, as more historical data becomes available, the estimates of AM may converge to a value other than $v(\pi_e)$. Thus, purely model-based estimators tend to have high bias, even asymptotically, as evidenced by the AM curve in Figure 1b. However, purely model-based methods also tend to have low variance because they do not contain any $\rho_t$ terms.

Between these two extremes lies a range of partial importance sampling estimators. Estimators that are close to the purely model-based estimators use $\rho_t$ terms only for small $t$, while estimators that are close to importance sampling estimators use $\rho_t$ terms with large $t$ approaching $L - 1$. Before formally defining one such partial importance sampling estimator, we present a few additional definitions. First, let $\text{IS}^{(j)}(D)$ denote an estimate of $\mathbf{E}[\sum_{t=0}^{j} \gamma^t R_t | H \sim \pi_e]$, constructed from $D$ using an importance sampling method like PDIS or WDR, which uses importance weights up to and including $\rho_t$. Similarly, let $\text{AM}^{(j)}(D)$ denote a primarily model-based prediction from $D$ of $\mathbf{E}[\sum_{t=j}^{\infty} \gamma^t R_t | H \sim \pi_e]$ that may not use $\rho_t$ terms with $t \geq j$.

We can now define a partial importance sampling estimator that we call the *off-policy j-step return*, $g^{(j)}(D)$, which uses an importance sampling based method to predict the outcome of using $\pi_e$ up until $R_j$ is generated, and the approximate model estimator to predict the outcomes thereafter. That is, for all $j \in \mathbb{N}_{\geq -1}$, let[8]

$$g^{(j)}(D) := \text{IS}^{(j)}(D) + \text{AM}^{(j+1)}(D)$$
$$g^{(\infty)}(D) := \lim_{j \to \infty} g^{(j)}(D). \quad (4)$$

We refer to $j$ as the *length* of the $j$-step return.

Notice that $g^{(-1)}(D)$ is a purely model-based estimator, $g^{(\infty)}(D)$ is an importance sampling estimator, and the other off-policy $j$-step returns are partial importance sampling estimators that blend between these two extremes. When $j$ is small, the off-policy $j$-step return is similar to AM, using importance sampling to predict only a few early rewards. When $j$ is large, it uses importance sampling to predict most of the rewards and the model only for a few rewards at the end of a trajectory. So, as $j$ increases, we expect the variance of the return to increase, but the bias to decrease.

We propose a new estimator, which we call the *blending IS and model* (BIM) estimator, that leverages this spectrum of estimators to blend together the IS and AM estimators in a way that minimizes MSE. It does this by computing a weighted average of the different length returns: $\text{BIM}(D) := \mathbf{x}^\intercal \mathbf{g}(D)$, where $\mathbf{x} := (x_{-1}, x_0, x_1, \ldots)^\intercal$ is an infinite-dimensional weight vector and $\mathbf{g}(D)$ is an infinite-dimensional vector of different length returns, $\mathbf{g}(D) := (g^{(-1)}(D), g^{(0)}(D), \ldots,)^\intercal$. Although in theory $\mathbf{x}$ can be infinite, in practice there is always finite data, so $\mathbf{x}$ is finite. The remaining question is then: how should we select the weights, $\mathbf{x}$?

A similar question has been studied before in reinforce-ment learning research when deciding how to weight $j$-step returns (not off-policy), as reviewed by Sutton & Barto (1998, Section 7.2). The most common solution, a complex return called the $\lambda$-*return*, uses $x_{-1} = 0$ and $x_j = (1-\lambda)\lambda^j$ for all other $j$. The $\lambda$-return is the foundation of the entire TD($\lambda$) family of algorithms, which includes the original linear-time algorithm (Sutton, 1988), least-squares formulations (Bradtke & Barto, 1996; Mahmood et al., 2014), methods for adapting $\lambda$ (Downey & Sanner, 2010), true-online methods (van Hasselt et al., 2014), and the recent emphatic methods (Mahmood et al., 2015).

---

[8] If prior knowledge about $d_0$ is available, then one might consider adding $g^{(-2)}(D)$ to denote the model's prediction of $v(\pi_e)$, which might differ from $g^{(-1)}(D)$.

Recent work has suggested that the $\lambda$-return could be replaced by more statistically principled complex returns like the $\gamma$-return (Konidaris et al., 2011) or $\Omega$-return (Thomas et al., 2015). For the finite-horizon setting and for $j \in \{0, \ldots, L-1\}$ the $\gamma$-return uses $x_j \coloneqq (\sum_{i=0}^{j} \gamma^{2i})^{-1} / \sum_{\hat{j}=0}^{L-1} (\sum_{i=0}^{\hat{j}} \gamma^{2i})^{-1}$, and the $\Omega$-return uses $x_j = \sum_{i=0}^{L-1} \Omega_n^{-1}(j,i) / \sum_{\hat{j},i=0}^{L-1} \Omega_n^{-1}(\hat{j},i)$, where $\Omega_n$ is the $L \times L$ covariance matrix where $\Omega_n(i,j) = \text{Cov}(g^{(i)}(D), g^{(j)}(D))$, and where both the $\gamma$ and $\Omega$-returns use $x_j = 0$ for $j \notin \{0, \ldots, L-1\}$.

The advantage of the $\gamma$-return over the $\lambda$-return is that it uses a more accurate model of how variance increases with the length of a return, which also eliminates the $\lambda$ hyperparameter used by the $\lambda$-return. The advantages of the $\Omega$-return over the $\gamma$-return are that it both uses a yet more-accurate estimate of how variance grows with the length of the return, which is computed from historical data, and that it better accounts for the fact that different length returns are *not* independent, i.e., $g^{(i)}(D)$ and $g^{(j)}(D)$ are not independent even if $i \neq j$.

However, none of these weighting schemes are sufficient for our needs because they do not cause BIM to necessarily be a strongly consistent estimator.[9] This is likely because they were all designed for the setting where only one trajectory is available, i.e., $n = 1$, while strong consistency is a property that deals with performance as $n \to \infty$. Furthermore, they were designed for *on-policy* policy evaluation.

We therefore propose a new weighting scheme (a new complex return for multiple trajectories) that directly optimizes our primary objective: the mean squared error. This new weighting scheme is $\mathbf{x}^\star \coloneqq \arg\min_{\mathbf{x} \in \mathbb{R}^\infty} \text{MSE}(\mathbf{x}^\intercal \mathbf{g}(D), v(\pi_e))$. Unfortunately, we typically cannot compute $\mathbf{x}^\star$, because we do not know $\text{MSE}(\mathbf{x}^\intercal \mathbf{g}(D), v(\pi_e))$ for any $\mathbf{x}$. Instead, we propose estimating $\mathbf{x}^\star$ by minimizing an approximation of $\text{MSE}(\mathbf{x}^\intercal \mathbf{g}(D), v(\pi_e))$. First, dealing with an infinite number of different return lengths is challenging. To avoid this, we propose only using a subset of the returns, $\{\mathbf{g}^{(j)}(D)\}$, for $j \in \mathcal{J}$, where $|\mathcal{J}| < \infty$. For all $j \notin \mathcal{J}$, we assign $\mathbf{x}_j = 0$. We suggest including $-1$ and $\infty$ in $\mathcal{J}$.

To simplify later notation, let $\mathbf{g}_\mathcal{J}(D) \in \mathbb{R}^{|\mathcal{J}|}$ be the elements of $\mathbf{g}(D)$ whose indexes are in $\mathcal{J}$—the returns that will not necessarily be given weights of zero. Also let $\mathcal{J}_j$ denote the $j^\text{th}$ element in $\mathcal{J}$. We can then estimate $\mathbf{x}^\star$ by:

$$\widehat{\mathbf{x}}^\star \in \arg\min_{\mathbf{x} \in \mathbb{R}^{|\mathcal{J}|}} \text{MSE}(\mathbf{x}^\intercal \mathbf{g}_\mathcal{J}(D), v(\pi_e)),$$

where our estimate of $x_j^\star$ is zero if $j \notin \mathcal{J}$ and our estimate of $x_{\mathcal{J}_j}^\star$ is $\widehat{x}_j^\star$ for $j \in \{1, \ldots, |\mathcal{J}|\}$.

Next, to avoid searching all of $\mathbb{R}^{|\mathcal{J}|}$ and also to serve as a form of regularization on $\widehat{\mathbf{x}}^\star$, we limit the set of $\mathbf{x}$ that we consider to the $|\mathcal{J}|$-simplex, i.e., we require $x_j \geq 0$ for all $j \in \{1, \ldots, |\mathcal{J}|\}$ and $\sum_{j=1}^{|\mathcal{J}|} x_j = 1$. We write $\Delta^{|\mathcal{J}|}$ to denote this set of weight vectors—the $|\mathcal{J}|$-simplex.

Using the bias-variance decomposition of MSE, we therefore have that

$$\widehat{\mathbf{x}}^\star \in \arg\min_{\mathbf{x} \in \Delta^{|\mathcal{J}|}} \text{Bias}(\mathbf{x}^\intercal \mathbf{g}_\mathcal{J}(D))^2 + \text{Var}(\mathbf{x}^\intercal \mathbf{g}_\mathcal{J}(D))$$
$$= \arg\min_{\mathbf{x} \in \Delta^{|\mathcal{J}|}} \mathbf{x}^\intercal [\Omega_n + \mathbf{b}_n \mathbf{b}_n^\intercal] \mathbf{x},$$

where $n$ remains the number of trajectories in $D$, $\Omega_n$ is the $|\mathcal{J}| \times |\mathcal{J}|$ covariance matrix where $\Omega_n(i,j) = \text{Cov}(\mathbf{g}^{(\mathcal{J}_i)}(D), \mathbf{g}^{(\mathcal{J}_j)}(D))$ and $\mathbf{b}_n$ is the $|\mathcal{J}|$-dimensional vector with $\mathbf{b}_n(j) = \mathbf{E}[\mathbf{g}^{(\mathcal{J}_j)}(D)] - v(\pi_e)$ for all $j \in \{1, \ldots, |\mathcal{J}|\}$.[10] This simplifies the problem of estimating the MSE for all possible $\mathbf{x}$ into estimating two terms: the bias vector, $\mathbf{b}_n$, and the covariance matrix, $\Omega_n$.

Let $\widehat{\mathbf{b}}_n$ and $\widehat{\Omega}_n$ be the estimates of $\mathbf{b}_n$ and $\Omega_n$ when there are $n$ trajectories in $D$. The exact scheme used to estimate $\mathbf{b}_n$ and $\Omega_n$ depends on the definitions of $\text{IS}^{(j)}(D)$ and $\text{AM}^{(j)}(D)$. In general, both terms are easier to estimate for unweighted importance sampling estimators like PDIS and DR than for weighted estimators like CWPDIS or WDR.

To make the dependence of BIM on the estimates of $\Omega_n$ and $\mathbf{b}_n$ explicit, and to summarize the approximations we have made, we redefine the BIM estimator as:

$$\text{BIM}(D, \widehat{\Omega}_n, \widehat{\mathbf{b}}_n) \coloneqq (\widehat{\mathbf{x}}^\star)^\intercal \mathbf{g}_\mathcal{J}(D),$$

where

$$\widehat{\mathbf{x}}^\star \in \arg\min_{\mathbf{x} \in \Delta^{|\mathcal{J}|}} \mathbf{x}^\intercal [\widehat{\Omega}_n + \widehat{\mathbf{b}}_n \widehat{\mathbf{b}}_n^\intercal] \mathbf{x}.$$

In the next section we propose using WDR as the importance sampling method, IS, and show how $\mathbf{b}_n$ and $\Omega_n$ can

---

[9] The $\lambda$-return with $\lambda = 1$ is defined to be $g^{(\infty)}(D)$ and is consistent, but it does not mix the two OPE methods at all.

[10] Since $\mathbf{b}_n$ (similarly, $\Omega_n$) already has a subscript, we write $\mathbf{b}_n(j)$ to denote the $j^\text{th}$ element of $\mathbf{b}_n$.

be approximated in this setting. First we show that if at least one of the returns included in $\mathcal{J}$ is a strongly consistent estimator of $v(\pi_e)$, and if the estimates of $\mathbf{b}_n$ and $\Omega_n$ are themselves strongly consistent, then BIM is a strongly consistent estimator of $v(\pi_e)$:

**Theorem 3.** *If Assumption 4 holds, there exists at least one $j \in \mathcal{J}$ such that $g^{(j)}(D)$ is a strongly consistent estimator of $v(\pi_e)$, $\widehat{\mathbf{b}}_n - \mathbf{b}_n \xrightarrow{a.s.} 0$, and $\widehat{\Omega}_n - \Omega_n \xrightarrow{a.s.} 0$, then $\text{BIM}(D, \widehat{\Omega}_n, \widehat{\mathbf{b}}_n) \xrightarrow{a.s.} v(\pi_e)$.* **Proof** *See Appendix E.*

## 8. Model and Guided Importance Sampling Combining (MAGIC) Estimator

In this section we propose using the BIM estimator with WDR as the importance sampling estimator. The resulting estimator combines purely model based estimates with the estimates of the guided importance sampling algorithm WDR, and so we call it the **m**odel **a**nd **g**uided **i**mportance sampling **c**ombining (MAGIC) estimator.

Although the derivation of how to properly define $\text{IS}^{(j)}(D)$ and $\text{AM}^{(j)}(D)$ in order to blend WDR with the approximate model is less obvious than one might expect and therefore an important technical detail, we relegate it to Appendix F due to space restrictions. The resulting definition of an off-policy $j$-step return is, for all $j \in \mathbb{N}_{\geq -1}$:

$$g^{(j)}(D) \coloneqq \underbrace{\sum_{i=1}^{n} \sum_{t=0}^{j} \gamma^t w_t^i R_t^{H_i}}_{(a)} + \underbrace{\sum_{i=1}^{n} \gamma^{j+1} w_j^i \hat{v}^{\pi_e}(S_{j+1}^{H_i})}_{(b)}$$
$$- \underbrace{\sum_{i=1}^{n} \sum_{t=0}^{j} \gamma^t \left( w_t^i \hat{q}^{\pi_e}\left(S_t^{H_i}, A_t^{H_i}\right) - w_{t-1}^i \hat{v}^{\pi_e}\left(S_t^{H_i}\right) \right)}_{(c)}.$$

where **(c)** is the combined control variate for both the importance sampling based term, **(a)**, and the model-based term, **(b)**, and where we use WDR's definition of $w_t^i$.

We estimate $\Omega_n$ from the $n$ trajectories in $D$ using a sample covariance matrix, $\widehat{\Omega}_n$. See Appendix G for details and pseudocode for MAGIC.

Estimating the bias vector, $\mathbf{b}_n$, is challenging because it has a strong dependence on the value that we wish we knew, $v(\pi_e)$. We cannot use AM's estimate as a stand-in for $v(\pi_e)$ because it would cause us to assume that AM's greatest weakness—its high bias—is negligible. We cannot use WDR's estimate (or any other importance sampling estimator's estimate) because our estimate of $\mathbf{b}_n$ would then conflate the high variance of importance sampling estimates with the bias that we wish to estimate.

When $n$, the number of trajectories in $D$, is small, variance tends to be the root cause of high MSE. We therefore propose using an estimate of $\mathbf{b}_n$ that is initially conservative—initially it underestimates the bias—but which becomes correct as $n$ increases. Let $\text{CI}(g^{(\infty)}(D), \delta)$ be a $1 - \delta$ confidence interval on the expected value of the random variable $g^{(\infty)}(D) = \text{WDR}(D)$. Intuitively, as $n$ increases we expect that this confidence interval will converge to $g^{(\infty)}(D)$, which in turn converges to $v(\pi_e)$. So, we estimate $\mathbf{b}_n(j)$, the bias of the off-policy $j$-step return, by its distance from the $10\%$ confidence interval. That is, we estimate $\mathbf{b}_n(j)$ as

$$\widehat{\mathbf{b}}_n(j) \coloneqq \text{dist}\left(g^{(\mathcal{J}_j)}(D), \text{CI}(g^{(\infty)}(D), 0.5)\right),$$

where $\text{dist}(y, \mathcal{Z})$ is the distance between $y \in \mathbb{R}$ and the set $\mathcal{Z} \subseteq \mathbb{R}$: $\text{dist}(y, \mathcal{Z}) = \min_{z \in \mathcal{Z}} |y - z|$. We use both the percentile bootstrap confidence interval (Efron & Tibshirani, 1993) and Chernoff-Hoeffding's inequality—whichever is tighter—for CI in our experiments.

In Theorem 4 we show that the MAGIC estimator is a strongly consistent estimator of $v(\pi_e)$ given one set of assumptions that we used to show that WDR is strongly consistent and that WDR is included as one of the off-policy $j$-step returns.

**Theorem 4** (MAGIC - strongly consistent). *If Assumptions 1 and 4 hold and $\infty \in \mathcal{J}$ then $\text{MAGIC}(D) \xrightarrow{a.s.} v(\pi_e)$.* **Proof** *See Appendix H.*

## 9. Empirical Studies (MAGIC)

Appendix I provides detailed experiments using MAGIC. In this section we provide an overview of these results. The first three plots in Figure 2 correspond to those in Figure 1, but include MAGIC. In general MAGIC does very well, tracking or exceeding the best performance of WDR and AM. However, in Figure 2c MAGIC does not perfectly track AM. The scale is logarithmic, so the difference between MAGIC and AM is small in comparison to the benefit of MAGIC over WDR. We hypothesize that the reason MAGIC does not match AM may be due to error in our estimates of $\Omega_n$ and $\mathbf{b}_n$.

Figure 2d is for an experimental setup that we call *Hybrid*, where early in trajectories there is partial observability (e.g., initial uncertainty about a student's knowledge in an intelligent tutoring system, or uncertainty about the state of the world in a robotic application). In these settings MAGIC outperforms all other estimators, even AM and WDR, by automatically leveraging WDR for the parts of trajectories where partial observability causes the model to be inaccurate, and AM for parts of the trajectories where the model is accurate. To emphasize this, we include MAGIC-B (B for *binary*) where $\mathcal{J} = \{-1, \infty\}$, so that BIM can only blend AM and WDR by placing weights on them. The relatively poor performance of MAGIC-B supports our use of off-policy $j$-step returns.

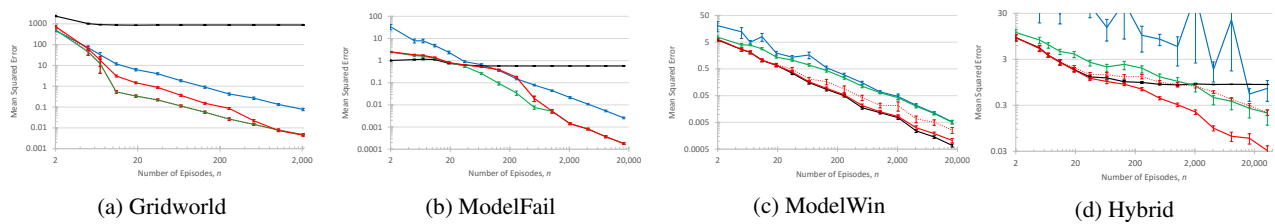

Figure 2: Empirical comparison of MAGIC to other estimators using the legend from Figure 1. All plots use the following legend (although only Figure 2d includes MAGIC-B):

— DR  — AM  — WDR  — MAGIC  ······ MAGIC-B

## 10. Conclusion

We have proposed several new OPE estimators and showed empirically that they outperform existing estimators. While previous OPE estimators that use importance sampling often failed to outperform the approximate model estimator (which does not use importance sampling), our new estimators often do, frequently by orders of magnitude. In cases where the approximate model estimator remains the best estimator, one of our new estimators, MAGIC, performs similarly. In other cases, MAGIC meets or exceeds the performance of state-of-the-art prior estimators.

## References

Bartle, Robert G. *The elements of integration and Lebesgue measure*. John Wiley & Sons, 2014.

Bhatnagar, S., Sutton, R. S., Ghavamzadeh, M., and Lee, M. Natural actor-critic algorithms. *Automatica*, 45(11): 2471–2482, 2009.

Bradtke, S.J. and Barto, A.G. Linear least-squares algorithms for temporal difference learning. *Machine Learning*, 22(1–3):33–57, March 1996.

Davison, A. C. and Hinkley, D. V. *Bootstrap Methods and their Application*. Cambridge University Press, Cambridge, 1997.

Downey, C. and Sanner, S. Temporal difference Bayesian model averaging: A Bayesian perspective on adapting lambda. In *Proceedings of the 27th International Conference on Machine Learning*, pp. 311–318, 2010.

Dudík, M., Langford, J., and Li, L. Doubly robust policy evaluation and learning. In *Proceedings of the Twenty-Eighth International Conference on Machine Learning*, pp. 1097–1104, 2011.

Efron, B. and Tibshirani, R. J. *An Introduction to the Bootstrap*. Chapman and Hall, London, 1993.

Hammersley, J. M. and Handscomb, D. C. Monte carlo methods, methuen & co. *Ltd., London*, pp. 40, 1964.

Heejung, H. and Robins, J. M. Doubly robust estimation in missing data and causal inference models. *Biometrics*, 61(4):962–973, 2005.

Jiang, N. and Li, L. Doubly robust off-policy evaluation for reinforcement learning. *ArXiv*, arXiv:1511.03722v1, 2015.

Konidaris, G. D., Niekum, S., and Thomas, P. S. $TD_\gamma$: Re-evaluating complex backups in temporal difference learning. In *Advances in Neural Information Processing Systems 24*, pp. 2402–2410. 2011.

Levine, S. and Koltun, V. Guided policy search. In *Proceedings of The 30th International Conference on Machine Learning*, pp. 1–9, 2013.

Mahmood, A. R., Hasselt, H., and Sutton, R. S. Weighted importance sampling for off-policy learning with linear function approximation. In *Advances in Neural Information Processing Systems 27*, 2014.

Mahmood, A. R., Yu, H., White, M., and Sutton, R. S. Emphatic temporal-difference learning. *ArXiv*, arXiv:1507.01569, 2015.

Mandel, T., Liu, Y., Levine, S., Brunskill, E., and Popović, Z. Offline policy evaluation across representations with applications to educational games. In *Proceedings of the 13th International Conference on Autonomous Agents and Multiagent Systems*, 2014.

Mandel, T., Liu, Y., Brunskill, E., and Popović, Z. Offline evaluation of online reinforcement learning algorithms. In *Proceedings of the Thirtieth Conference on Artificial Intelligence*, 2016.

Mittelhammer, R. C. *Mathematical statistics for economics and business*, volume 78. Springer, 1996.

Powell, M. J. D. and Swann, J. Weighted uniform sampling: a Monte Carlo technique for reducing variance. *Journal of the Institute of Mathematics and its Applications*, 2(3):228–236, 1966.

# A. Preliminaries

In this section we present additional notation, definitions, properties, (known) theorems, corollaries, and lemmas that are useful when we prove theorems later.

Let $H^t \coloneqq (S_0, A_0, R_0, S_1, \ldots, S_{t-1}, A_{t-1}, R_{t-1}, S_t)$ be the first $t$ transitions in the episode $H$. We call $H^t$ a *partial trajectory* of length $t$. Notice that we use subscripts on trajectories to denote the trajectory's index in $D$ and superscripts to denote partial trajectories—$H_i^t$ is the first $t$ transitions of the $i^{\text{th}}$ trajectory in $D$. Let $\mathcal{H}^t$ be the set of all possible partial trajectories of length $t$.

For all $(\pi, s) \in \Pi \times \mathcal{S}$, let $\text{supp}_s(\pi)$ be the set of actions that have non-zero probability when the policy $\pi$ is used to select an action in state $s$, i.e., $\text{supp}_s(\pi) \coloneqq \{a \in \mathcal{A} : \pi(a|s) \neq 0\}$. Similarly, let $\text{supp}(\pi, t) \coloneqq \{h^t \in \mathcal{H}^t : \Pr(H^t = h^t|\pi) \neq 0\}$.

Later we will need to bound terms like $\rho_t^i R_t^i$ for some $t$ and $i$. Notice that even if $\rho_t^i < \beta$, it is possible for $\rho_t^i R_t^i > \beta r_{\max}$ if $r_{\max}$ is negative, since $\rho_t^i$ could be zero. Additionally, sometimes we may deal with $r_{\max}$ terms and other times $r_{\max}^{\text{model}}$. To avoid explicitly handling these cases, we will bound terms using loose bounds that depend on a new term: $r_{\max}^\star \coloneqq \max\{|r_{\min}|, |r_{\max}|, |r_{\min}^{\text{model}}|, |r_{\max}^{\text{model}}|\}$.

**Definition 1** (Almost Sure Convergence). *A sequence of random variables, $(X_n)_{n=1}^\infty$, converges almost surely to the random variable $X$ if*

$$\Pr\left(\lim_{n \to \infty} X_n = X\right) = 1.$$

We write $X_n \xrightarrow{\text{a.s.}} X$ to denote that the sequence $(X_n)_{n=1}^\infty$ convergences almost surely to $X$.

**Definition 2.** *Let $\theta$ be a real number and $(\hat{\theta}_n)_{n=1}^\infty$ be an infinite sequence of random variables. We call $\hat{\theta}_n$, a (strongly) **consistent estimator** of $\theta$ if and only if $\hat{\theta}_n \xrightarrow{\text{a.s.}} \theta$.*

Notice that an estimator being unbiased does *not* mean that it is also strongly consistent—estimators can be any combination of biased/unbiased and consistent/inconsistent. Next we present several known properties of almost sure convergence (Mittelhammer, 1996, Section 5.5).

**Property 1.** *[Continuous mapping theorem] $X_n \xrightarrow{\text{a.s.}} X$ implies that $f(X_n) \xrightarrow{\text{a.s.}} f(X)$ for every continuous function $f$.*

**Property 2.** *Let $X_n$ and $Y_n$ be sequences of random variables and $X$ and $Y$ be random variables. If $X_n \xrightarrow{\text{a.s.}} X$, $Y_n \xrightarrow{\text{a.s.}} Y$, and if $\Pr(Y = 0) = 0$, then $\frac{X_n}{Y_n} \xrightarrow{\text{a.s.}} \frac{X}{Y}$.*

**Property 3.** *If $\{X_n^i\}_{i=1}^m$ are $m < \infty$ sequences of random variables such that $X_n^i \xrightarrow{\text{a.s.}} X^i$ for all $i \in \{1, \ldots, m\}$, then $\sum_{i=1}^m X_n^i \xrightarrow{\text{a.s.}} \sum_{i=1}^m X^i$.*

We will require an additional property of almost sure convergence that is similar to Property 3, but which allows for the sum over a countably infinite number of sequences of random variables, i.e., $m = \infty$. In order to establish this property we begin with Lebesgue's dominated convergence theorem:

**Theorem 5** (Lebesgue's Dominated Convergence Theorem). *Let $(f_n)_{n=1}^\infty$ be a sequence of integrable functions that converges almost everywhere to a real-valued measurable function $f$. If there exists an integrable function[11] $g$ such that $|f_n| \leq g$ for all $n$, then*

$$\lim_{n \to \infty} \int f_n \, d\mu = \int f \, d\mu.$$

*Proof.* See the work of (Bartle, 2014, Theorem 5.6). $\square$

Next we use Lebesgue's dominated convergence theorem to show conditions under which we can reverse the order of a limit and an infinite summation:

**Lemma 1.** *Let $\{x_n^i\}_{i=0}^\infty$ be a countably infinite number of real-valued sequences indexed by $i$, such that $\lim_{n \to \infty} x_n^i = x^i$ for all $i \in \mathbb{N}_{\geq 0}$. If there exists a function $g : \mathbb{N}_{\geq 0} \to \mathbb{R}$ such that $|x_n^i| \leq g(i)$ for all $n \in \mathbb{N}_{>0}$ and $i \in \mathbb{N}_{\geq 0}$, and $\sum_{i=0}^\infty g(i) < \infty$, then*

$$\lim_{n \to \infty} \sum_{i=0}^\infty x_n^i = \sum_{i=0}^\infty \lim_{n \to \infty} x_n^i.$$

*Proof.* We apply Lebesgue's dominated convergence theorem (Theorem 5), where for all $(n, i) \in \mathbb{N}_{>0} \times \mathbb{N}_{\geq 0}$, $f_n(i) = X_n^i$, $f(i) = x^i$, and $\mu$ is the counting measure on the measure space $(\mathbb{N}_{\geq 0}, \mathcal{P}(\mathbb{N}_{\geq 0}))$, where $\mathcal{P}(\mathbb{N}_{\geq 0})$ is the power set of $\mathbb{N}_{\geq 0}$. $\square$

We can now establish our desired property about almost sure convergence:

**Property 4.** *Let $\{X_n^i\}_{i=0}^\infty$ be a countably infinite number of sequences of random variables such that $X_n^i \xrightarrow{\text{a.s.}} X^i$ for all $i \in \mathbb{N}_{\geq 0}$. If there exists a function $g : \mathbb{N}_{\geq 0} \to \mathbb{R}$ such that $|X_n^i| \leq g(i)$ surely for all $(n, i) \in \mathbb{N}_{>0} \times \mathbb{N}_{\geq 0}$, and $\sum_{i=0}^\infty g(i) < \infty$, then $\sum_{i=0}^\infty X_n^i \xrightarrow{\text{a.s.}} \sum_{i=0}^\infty X^i$.*

---

[11]To conform to standard notations elsewhere, here we reuse the symbol $g$, which was previously used to denote the return of a trajectory, $g(H)$. The two uses of $g$ are sufficiently dissimilar that this reuse should not cause confusion.

*Proof.*

$$\Pr\left(\lim_{n\to\infty}\sum_{i=0}^{\infty} X_n^i = \sum_{i=0}^{\infty} X^i\right)$$

$$\overset{(a)}{\geq} \Pr\left(\bigcap_{i=0}^{\infty}\left(\lim_{n\to\infty} X_n^i = X^i\right) \cap \underbrace{\left(\sum_{i=0}^{\infty}\lim_{n\to\infty} X_n^i = \sum_{i=0}^{\infty} X^i\right)}_{(b)}\right)$$

$$\overset{(c)}{\geq} \Pr\left(\bigcap_{i=0}^{\infty}\left(\lim_{n\to\infty} X_n^i = X^i\right) \cap \underbrace{\left(\bigcap_{i=0}^{\infty}\left(\lim_{n\to\infty} X_n^i = X^i\right)\right)}_{(d)}\right)$$

$$= \Pr\left(\bigcap_{i=0}^{\infty}\left(\lim_{n\to\infty} X_n^i = X^i\right)\right)$$

$$= 1 - \Pr\left(\underbrace{\bigcup_{i=0}^{\infty}\left(\lim_{n\to\infty} X_n^i \neq X^i\right)}_{(e)}\right)$$

$$= 1,$$

where **(a)** comes from Lemma 1 which ensures that

$$\bigcap_{i=0}^{\infty}\left(\lim_{n\to\infty} X_n^i = X^i\right) \implies \left(\lim_{n\to\infty}\sum_{i=0}^{\infty} X_n^i = \sum_{i=0}^{\infty}\lim_{n\to\infty} X_n^i\right),$$

**(c)** holds because **(d)** $\implies$ **(b)**, and **(e)** has zero measure because it is the countable union of zero measure sets by the assumption that $X_n^i \xrightarrow{a.s.} X^i$ for all $i \in \mathbb{N}_{\geq 0}$. $\square$

Next we show that if a sequence of random variables, $X_n$, converges almost surely to a random variable, $X$, then the expected value of $X_n$ converges to the expected value of $X$.

**Lemma 2.** *If $(X_i)_{i=1}^{\infty}$ is a sequence of uniformly bounded real-valued random variables and if $X_n \xrightarrow{a.s.} X$, then $\lim_{n\to\infty} \mathbf{E}[X_n] = \mathbf{E}[X]$.*

*Proof.* Let $X_n$ (for all $n$) and $X$ be random variables on the probability space $(\Omega, \Sigma, P)$ and let $\mathcal{A} = \{\omega \in \Omega : \lim_{n\to\infty} X_n = X\}$. Then:

$$\lim_{n\to\infty} \mathbf{E}[X_n] = \lim_{n\to\infty} \int_{\Omega} X_n \mathrm{d}P$$

$$\overset{(a)}{=} \int_{\Omega} \lim_{n\to\infty} X_n \mathrm{d}P$$

$$= \underbrace{\int_{\mathcal{A}} \lim_{n\to\infty} X_n \mathrm{d}P}_{(b)} + \underbrace{\int_{\Omega\setminus\mathcal{A}} \lim_{n\to\infty} X_n \mathrm{d}P}_{(c)},$$

where **(a)** comes from the bounded convergence theorem. For term **(b)**, notice that for all $\omega \in \mathcal{A}$, $\lim_{n\to\infty} X_n = X$. For term **(c)**, notice that by the assumption that $X_n \xrightarrow{a.s.} X$, we have that $\Omega \setminus \mathcal{A}$ has measure zero. So:

$$\lim_{n\to\infty} \mathbf{E}[X_n] = \int_{\mathcal{A}} X \mathrm{d}P$$

$$= \int_{\mathcal{A}} X \mathrm{d}P + \int_{\Omega\setminus\mathcal{A}} X \mathrm{d}P$$

$$= \mathbf{E}[X].$$

$\square$

Next we present a lemma that relates almost sure convergence of estimators to mean squared error. Let $\hat{\theta}$ be an estimator of $\theta$. Recall that:

$$\mathrm{MSE}(\hat{\theta}, \theta) := \mathbf{E}\left[(\hat{\theta} - \theta)^2\right].$$

We show that a sequence, $(X_n)_{n=1}^{\infty}$ converges almost surely to $X$ if and only if $\lim_{n\to\infty} \mathrm{MSE}(X_n, X) = 0$.

**Lemma 3.** *If $(X_i)_{i=1}^{\infty}$ is a sequence of uniformly bounded real-valued random variables, then $X_n \xrightarrow{a.s.} X$ if and only if $\lim_{n\to\infty} \mathrm{MSE}(X_n, X) = 0$.*

*Proof.* We show each direction separately. First we show that $X_n \xrightarrow{a.s.} X$ implies $\lim_{n\to\infty} \mathrm{MSE}(X_n, X) = 0$.

$$\mathrm{MSE}(X_n, X) = \mathbf{E}[(X_n - X)^2]$$
$$= \mathbf{E}[Y_n],$$

where $Y_n := (X_n - X)^2$. By the continuous mapping theorem we have that $Y_n \xrightarrow{a.s.} (X - X)^2 = 0$. So, by Lemma 2 (applied to $\mathbf{E}[Y_n]$) we have that

$$\lim_{n\to\infty} \mathrm{MSE}(X_n, X) = \mathbf{E}[0]$$
$$= 0.$$

Next we show the other direction: that $\lim_{n\to\infty} \mathrm{MSE}(X_n, X) = 0$ implies $X_n \xrightarrow{a.s.} X$. Let $X$ and all $X_n$ be random variables on the probability space $(\Omega, \Sigma, P)$, $\mathcal{A} = \{\omega \in \Omega : \lim_{n\to\infty} \mathrm{MSE}(X_n, X) = 0\}$, and $\mathcal{B} = \{\omega \in \mathcal{A} : \lim_{n\to\infty} X_n \neq X\}$. If $\lim_{n\to\infty} \mathrm{MSE}(X_n, X) = 0$, then by the definition of

MSE we have that:

$$\begin{aligned}
0 &= \lim_{n\to\infty} \int_\Omega (X_n - X)^2 \, dP \\
&\stackrel{(a)}{=} \int_\Omega \left(\lim_{n\to\infty} X_n - X\right)^2 dP \\
&= \underbrace{\int_\mathcal{B} \left(\lim_{n\to\infty} X_n - X\right)^2 dP}_{(b)} \\
&\quad + \underbrace{\int_{\mathcal{A}\setminus\mathcal{B}} \left(\lim_{n\to\infty} X_n - X\right)^2 dP}_{(c)} \\
&\quad + \underbrace{\int_{\Omega\setminus\mathcal{A}} \left(\lim_{n\to\infty} X_n - X\right)^2 dP}_{(d)},
\end{aligned}$$

where we get **(a)** by using the bounded convergence theorem to pass the limit inside the integral and the fact that $(X_n - X)^2$ is a continuous function of $X_n$ to then move the limit to the $X_n$ term. Notice that **(b)**, **(c)**, and **(d)** are all positive, and so they must all be zero for the equality with zero to hold. We have that **(d)** is necessarily zero due to the definition of $\mathcal{A}$ and our assumption that $\lim_{n\to\infty} \mathrm{MSE}(X_n, X) = 0$. Similarly, **(c)** is zero because, from the definition of $\mathcal{B}$, $\mathcal{A}\setminus\mathcal{B}$ causes $\lim_{n\to\infty} X_n = X$. However, in **(b)**, by the definition of $\mathcal{B}$, $\lim_{n\to\infty} X_n - X$ is non-zero, and so for the equality with zero to hold, $\mathcal{B}$ must have measure zero. That is, $\Pr(\lim_{n\to\infty} X_n \neq X) = 0$, and thus $\Pr(\lim_{n\to\infty} X_n = X) = 1$. $\square$

Next we show that if two sequences of random variables converge to the same random variable, then any sequence of random variables bounded between the two sequences must also converge to the same random variable.

**Lemma 4.** *If $X_n \xrightarrow{a.s.} X$, $Z_n \xrightarrow{a.s.} X$, and for all $n$, $X_n \leq Y_n \leq Z_n$, then $Y_n \xrightarrow{a.s.} X$.*

*Proof.*

$$\Pr\left(\lim_{n\to\infty} Y_n = X\right) = \Pr\left(\left(\lim_{n\to\infty} Y_n \leq X\right) \cap \left(\lim_{n\to\infty} Y_n \geq X\right)\right) \quad (5)$$

Since

$$\Pr\left(\lim_{n\to\infty} Y_n \geq X\right) \geq \Pr\left(\lim_{n\to\infty} X_n \geq X\right)$$
$$\geq \Pr\left(\lim_{n\to\infty} X_n = X\right)$$
$$= 1,$$

and

$$\Pr\left(\lim_{n\to\infty} Y_n \leq X\right) \geq \Pr\left(\lim_{n\to\infty} Z_n \leq X\right)$$
$$\geq \Pr\left(\lim_{n\to\infty} Z_n = X\right)$$
$$= 1,$$

we have that (5) is the probability of the joint occurance of two probability one events, and so

$$\Pr\left(\lim_{n\to\infty} Y_n = X\right) = 1.$$

$\square$

Next we show that if the difference between two sequences converges almost surely to zero, then we can substitute one sequence for the other as an input to a continuous function without changing the almost sure convergence properties of the function:

**Lemma 5.** *If $f$ is a continuous function, $f(X_n) \xrightarrow{a.s.} X$, and $Y_n - X_n \xrightarrow{a.s.} 0$, then $f(Y_n) \xrightarrow{a.s.} X$.*

*Proof.*

$$\Pr\left(\lim_{n\to\infty} f(Y_n) = X\right) = \Pr\left(\lim_{n\to\infty} f(Y_n - X_n + X_n) = X\right)$$

$$\stackrel{(a)}{=} \Pr\left(f\left(\lim_{n\to\infty} Y_n - X_n + X_n\right) = X\right)$$

$$\stackrel{(b)}{\geq} \Pr\left(\left(\lim_{n\to\infty} Y_n - X_n = 0\right) \right.$$
$$\left. \cap \left(f\left(\lim_{n\to\infty} X_n\right) = X\right)\right)$$

$$= \Pr\left(\left(\lim_{n\to\infty} Y_n - X_n = 0\right) \right.$$
$$\left. \cap \left(\lim_{n\to\infty} f(X_n) = X\right)\right)$$

$$\stackrel{(c)}{=} 1,$$

where **(a)** holds because $f$ is a continuous function, and where **(b)** holds because it gives sufficient conditions for the event in the line above to hold, and **(c)** holds because under our assumptions the two events both occur with probability one. So we can conclude that $f(Y_n) \xrightarrow{a.s.} X$. $\square$

Next we review two standard forms of the strong law of large numbers.

**Theorem 6** (Khintchine Strong Law of Large Numbers). *Let $\{X_i\}_{i=1}^\infty$ be independent and identically distributed random variables. Then $(\frac{1}{n}\sum_{i=1}^n X_i)_{n=1}^\infty$ is a sequence of random variables that converges almost surely to $\mathbf{E}[X_1]$.*

*Proof.* See the work of Sen & Singer (1993, Theorem 2.3.13). □

**Theorem 7** (Kolmogorov Strong Law of Large Numbers). *Let $\{X_i\}_{i=1}^\infty$ be independent (not necessarily identically distributed) random variables. If all $X_i$ have the same mean and bounded variance (i.e., there is a finite constant $b$ such that for all $i \geq 1$, $\mathrm{Var}(X_i) \leq b$), then $(\frac{1}{n}\sum_{i=1}^n X_i)_{n=1}^\infty$ is a sequence of random variables that converges almost surely to $\mathbf{E}[X_1]$.*

*Proof.* See the work of Sen & Singer (1993, Theorem 2.3.10 with Proposition 2.3.10). □

In Corollary 1 we present a simple extension of Kolmogorov's strong law of large numbers that we often still refer to as Kolmogorov's strong law of large numbers:

**Corollary 1.** *Let $\{X_i\}_{i=1}^\infty$ be independent (not necessarily identically distributed) random variables. If all $X_i$ have the same mean and are uniformly bounded by a finite constant $b$, then $(\frac{1}{n}\sum_{i=1}^n X_i)_{n=1}^\infty$ is a sequence of random variables that converges almost surely to $\mathbf{E}[X_1]$.*

*Proof.* For all $i \in \mathbb{N}_{>0}$ we have that $|X_i| \leq b$ surely, so from Popoviciu's inequality, $\mathrm{Var}(X_i) \leq b^2$, and so we can apply Theorem 7. □

We now turn to results that are more specific to reinforcement learning and off-policy policy evaluation. Lemma 6 establishes a relationship between the expected values of $\hat{r}^{\pi_e}(s,i)$ and $\hat{r}^{\pi_e}(s,A,i)$ for all $i$ if $A$ is generated by some policy $\pi$.

**Lemma 6.** *Let $(\pi_e, \pi) \in \Pi^2$, where $(\pi(a|s) = 0) \implies (\pi_e(a|s) = 0)$ for all $(a,s) \in \mathcal{A} \times \mathcal{S}$. Then for all $(s,i) \in \mathcal{S} \times \mathbb{N}_{\geq 0}$,*

$$\hat{r}^{\pi_e}(s,i) = \mathbf{E}\left[\frac{\pi_e(A|s)}{\pi(A|s)}\hat{r}^{\pi_e}(s,A,i)\bigg| A \sim \pi\right].$$

*Proof.* First, recall from (1) that for all $(s,i) \in \mathcal{S} \times \{1,\ldots,n\}$:

$$\hat{r}^{\pi_e}(s,i) := \sum_{a \in \mathcal{A}} \pi_e(a|s)\hat{r}^{\pi_e}(s,a,i)$$

$$= \sum_{a \in \mathrm{supp}_s(\pi_e)} \pi_e(a|s)\hat{r}^{\pi_e}(s,a,i)$$

$$\stackrel{(a)}{=} \sum_{a \in \mathrm{supp}_s(\pi)} \pi_e(a|s)\hat{r}^{\pi_e}(s,a,i)$$

$$= \sum_{a \in \mathrm{supp}_s(\pi)} \frac{\pi(a|s)}{\pi(a|s)}\pi_e(a|s)\hat{r}^{\pi_e}(s,a,i)$$

$$= \sum_{a \in \mathrm{supp}_s(\pi)} \pi(a|s)\frac{\pi_e(a|s)}{\pi(a|s)}\hat{r}^{\pi_e}(s,a,i)$$

$$= \mathbf{E}\left[\frac{\pi_e(A|s)}{\pi(A|s)}\hat{r}^{\pi_e}(s,A,i)\bigg| A \sim \pi\right].$$

where **(a)** holds by the assumption that $(\pi(a|s) = 0) \implies (\pi_e(a|s) = 0)$ for all $(a,s) \in \mathcal{A} \times \mathcal{S}$. □

Corollary 2 extends Lemma 6 to show a relationship between $\hat{v}^{\pi_e}(s)$ and the expected value of $\hat{q}^{\pi_e}(s,A,i)$ if $A$ is generated by some policy $\pi$:

**Corollary 2.** *Let $(\pi_e, \pi) \in \Pi^2$, where $(\pi(a|s) = 0) \implies (\pi_e(a|s) = 0)$ for all $(a,s) \in \mathcal{A} \times \mathcal{S}$. Then for all $s \in \mathcal{S}$,*

$$\hat{v}^{\pi_e}(s) = \mathbf{E}\left[\frac{\pi_e(A|s)}{\pi(A|s)}\hat{q}^{\pi_e}(s,A)\bigg| A \sim \pi\right].$$

*Proof.* We have from Lemma 6 that for all $i \in \mathbb{N}_{\geq 0}$,

$$\hat{r}^{\pi_e}(s,i) = \mathbf{E}\left[\frac{\pi_e(A|s)}{\pi(A|s)}\hat{r}^{\pi_e}(s,A,i)\bigg| A \sim \pi\right].$$

Summing both sides over $t$ and multiplying by $\gamma^t$ we have that:

$$\underbrace{\sum_{t=0}^\infty \gamma^t \hat{r}^{\pi_e}(s,t)}_{=\hat{v}^{\pi_e}(s)} = \sum_{t=0}^\infty \gamma^t \mathbf{E}\left[\frac{\pi_e(A|s)}{\pi(A|s)}\hat{r}^{\pi_e}(s,A,t)\bigg| A \sim \pi\right]$$

$$\hat{v}^{\pi_e}(s) = \mathbf{E}\left[\frac{\pi_e(A|s)}{\pi(A|s)}\underbrace{\sum_{t=0}^\infty \gamma^t \hat{r}^{\pi_e}(s,A,t)}_{=\hat{q}^{\pi_e}(s,A)}\bigg| A \sim \pi\right]$$

$$= \mathbf{E}\left[\frac{\pi_e(A|s)}{\pi(A|s)}\hat{q}^{\pi_e}(s,A)\bigg| A \sim \pi\right].$$

□

Before presenting the next theorem, notice that we can express the DR estimator, (2), as $\mathrm{DR}(D) = \frac{1}{n}\sum_{i=1}^n \mathrm{DR}_i(D)$

if

$$\mathrm{DR}_i(D) := \sum_{t=0}^{\infty} \gamma^t \rho_t^i R_t^{H_i}$$
$$- \sum_{t=0}^{\infty} \gamma^t \left( \rho_t^i \hat{q}^{\pi_e} \left( S_t^{H_i}, A_t^{H_i} \right) - \rho_{t-1}^i \hat{v}^{\pi_e} \left( S_t^{H_i} \right) \right).$$

Lemma 7 gives conditions under which the DR estimator is an unbiased estimator of $v(\pi_e)$ when using only one trajectory. This lemma is the bulk of the proof that the full DR estimator is unbiased—we have placed it in a separate lemma because it is also a useful result when showing that the DR estimator is strongly consistent.

**Lemma 7.** *If Assumption 1 holds then $\mathbf{E}[\mathrm{DR}_i(D)] = v(\pi_e)$ for all $i \in \{1, \ldots, n\}$.*

*Proof.* Recall that

$$\mathrm{DR}_i(D) := \sum_{t=0}^{\infty} \gamma^t \rho_t^i R_t^{H_i}$$
$$- \sum_{t=0}^{\infty} \gamma^t \left( \rho_t^i \hat{q}^{\pi_e} \left( S_t^{H_i}, A_t^{H_i} \right) - \rho_{t-1}^i \hat{v}^{\pi_e} \left( S_t^{H_i} \right) \right).$$

First, notice that $\sum_{t=0}^{\infty} \gamma^t \rho_t^{H_i} R_t^{H_i}$ is the *per-decision importance sampling* (PDIS) estimator, which is known to be an unbiased estimator of $v(\pi_e)$ (Precup et al., 2000; Thomas, 2015b). So, we need only show that the remaining terms in the definition of $\mathrm{DR}_i(D)$ have expected value zero, i.e., that

$$\mathbf{E}\left[\sum_{t=0}^{\infty} \gamma^t \rho_t^i \hat{q}^{\pi_e}\left(S_t^{H_i}, A_t^{H_i}\right)\right] = \mathbf{E}\left[\sum_{t=0}^{\infty} \gamma^t \rho_{t-1}^i \hat{v}^{\pi_e}\left(S_t^{H_i}\right)\right].$$

By Corollary 2 (which requires Assumption 1) we have that

$$\mathbf{E}\left[\sum_{t=0}^{\infty} \gamma^t \rho_{t-1}^i \hat{v}^{\pi_e}\left(S_t^{H_i}\right)\right]$$
$$= \mathbf{E}\left[\sum_{t=0}^{\infty} \gamma^t \rho_{t-1}^i \frac{\pi_e\left(A_t^{H_i}|S_t^{H_i}\right)}{\pi_i\left(A_t^{H_i}|S_t^{H_i}\right)} \hat{q}^{\pi_e}\left(S_t^{H_i}, A_t^{H_i}\right)\right]$$
$$= \mathbf{E}\left[\sum_{t=0}^{\infty} \gamma^t \rho_t^i \hat{q}^{\pi_e}\left(S_t^{H_i}, A_t^{H_i}\right)\right].$$

$\square$

For completeness, next we show formally the obvious result that Assumption 1 implies that partial trajectories that occur under the evaluation policy must occur under the behavior policy.

**Lemma 8.** *Assumption 1 implies that if $\Pr(H^t = h^t|\pi_i) = 0$, then $\Pr(H^t = h^t|\pi_e) = 0$ for all $i \in \{1, \ldots, n\}$, $h^t := (s_0, a_0, r_0, s_1, \ldots, s_{t-1}, a_{t-1}, r_{t-1}, s_t) \in \mathcal{H}^t$, and $0 \le t < \infty$.*

*Proof.* If $t = 0$ then $h^t = (s_0)$, which does not depend on the policy, so clearly if $\Pr(H^0 = h^0|\pi_i) = 0$ then $\Pr(H^0 = h^0|\pi_e) = 0$. Hereafter we assume $1 \le t < \infty$. Notice that for any $\pi \in \Pi$,

$$\Pr(H^t = h^t|\pi)$$
$$\stackrel{(a)}{=} \Pr(S_0 = s_0) \Pr(A_0 = a_0|S_0 = s_0, \pi)$$
$$\times \Big( \prod_{i=1}^{t-1} \Pr(S_i = s_i|S_{i-1} = s_{i-1}, A_{i-1} = a_{i-1})$$
$$\times \Pr(R_{i-1} = r_{i-1}|S_{i-1} = s_{i-1}, A_{i-1} = a_{i-1}, S_i = s_i)$$
$$\times \Pr(A_i = a_i|S_i = s_i, \pi) \Big)$$
$$\times \Pr(S_t = s_t|S_{t-1} = s_{t-1}, A_{t-1} = a_{t-1})$$
$$\times \Pr(R_{t-1} = r_{t-1}|S_{t-1} = s_{t-1}, A_{t-1} = a_{t-1}, S_t = s_t)$$
$$\stackrel{(b)}{=} d_0(s_0)\pi(a_0|s_0)P(s_t|s_{t-1}, a_{t-1})R(r_{t-1}|s_{t-1}, a_{t-1}, s_t)$$
$$\times \prod_{i=1}^{t-1} P(s_i|s_{i-1}a_{i-1})R(r_{i-1}|s_{i-1}, a_{i-1}, s_i)\pi(a_i|s_i).$$

where **(a)** comes from repeated application of the rule that, for any random variables $X$ and $Y$, $\Pr(X = x, Y = y) = \Pr(X = x) \Pr(Y = y|X = x)$ and the Markov property for state transitions, actions, and rewards, and **(b)** comes from the definitions of $d_0, \pi, R$ and $P$ in MDPNv1.

So, if $\Pr(H_t = h_t|\pi_i) = 0$, then one of the terms in the product above (using $\pi_i$ for $\pi$) must be zero. If that term is not a $\pi_i$ term, then it also shows up in $\Pr(H_t = h_t|\pi_e)$, and so $\Pr(H_t = h_t|\pi_e) = 0$. If the term is a $\pi_i$ term, then by Assumption 1, the corresponding $\pi_e$ term must also be zero, and so $\Pr(H_t = h_t|\pi_i) = 0$. $\square$

Next, recall the known result that the ratio of partial trajectory probabilities under two different policies can be written in terms of the two policies:

**Lemma 9.** *Let $\pi_e$ and $\pi_b$ be any two policies and $t \in \mathbb{N}_{>0}$. Let $h_t$ be any history of length $t$ that has non-zero probability under $\pi_b$, i.e., $\Pr(H_t = h_t|\pi_b) \ne 0$. Then*

$$\frac{\Pr(H_t = h_t|\pi_e)}{\Pr(H_t = h_t|\pi_b)} = \prod_{i=0}^{t-1} \frac{\pi_e(a_i|s_i)}{\pi_b(a_i|s_i)}.$$

*Proof.* See the works of (Precup et al., 2000) or (Thomas, 2015b, Lemma 1). $\square$

Next we establish Lemma 10, which states that we can use importance sampling to generate unbiased estimates of any function of partial trajectories in $D$. Recall that whenever we write $H_i$ (or $H_i^t$) we always mean a trajectory generated by $\pi_i$, so $H_i \sim \pi_i$.

**Lemma 10.** *If Assumption* 1 *holds, then for all* $(t, i) \in \mathbb{N}_{\geq -1} \times \{1, \ldots, n\}$:

$$\mathbf{E}[\rho_t^i f(H_i^{t+1})] = \mathbf{E}[f(H^{t+1})|H^{t+1} \sim \pi_e],$$

*for any real-valued function $f$.*

*Proof.* If $t = -1$ then $H^{t-1} = (S_0)$, which does not depend on the policy, so the result is immediate. If $t \geq 0$:

$$\mathbf{E}[\rho_t^i f(H_i^{t+1})] = \mathbf{E}\left[\prod_{j=0}^t \frac{\pi_e\left(A_j^{H_i}|S_j^{H_i}\right)}{\pi_i\left(A_j^{H_i}|S_j^{H_i}\right)} f(H_i^{t+1})\right]$$

$$\stackrel{(a)}{=} \mathbf{E}\left[\frac{\Pr\left(H_i^{t+1} = h_i^{t+1}|\pi_e\right)}{\Pr\left(H_i^{t+1} = h_i^{t+1}|\pi_i\right)} f(H_i^{t+1})\right]$$

$$= \sum_{\text{supp}(\pi_i, t+1)} \Pr\left(H^{t+1} = h^{t+1}|\pi_i\right)$$

$$\times \frac{\Pr\left(H^{t+1} = h^{t+1}|\pi_e\right)}{\Pr\left(H^{t+1} = h^{t+1}|\pi_i\right)} f(H^{t+1})$$

$$= \sum_{\text{supp}(\pi_i, t+1)} \Pr\left(H^{t+1} = h^{t+1}|\pi_e\right) f(H^{t+1})$$

$$\stackrel{(b)}{=} \sum_{\text{supp}(\pi_e, t+1)} \Pr\left(H^{t+1} = h^{t+1}|\pi_e\right) f(H^{t+1})$$

$$= \mathbf{E}[f(H^{t+1})|H^{t+1} \sim \pi_e],$$

where **(a)** comes from Lemma 9 and **(b)** comes from Lemma 8, which requires Assumption 1. □

We can use Lemma 10 to show the well-known result that the expected value of an importance weight is one:

**Lemma 11.** *For all $\pi_i$ and $t \in \mathbb{N}_{\geq -1}$, if Assumption* 1 *holds, then $\mathbf{E}[\rho_t^i] = 1$.*

*Proof.* This follows from Lemma 10 with $f(H^{t+1}) := 1$. □

Next we establish a lemma that will be crucial to showing that the WDR estimator is strongly consistent:

**Lemma 12.** *For all $t \in \mathbb{N}_{\geq 0}$, let $f_t : \mathcal{H}^{t+1} \to \mathbb{R}$. If Assumption* 1 *holds, $f_t = 0$ for all $t \in \mathbb{N}_{\geq L}$, and either:*

- **Case 1:** *Assumptions* 2 *and* 3 *hold.*

  *or*

- **Case 2:** *Assumption* 4 *holds and there is a finite $f_{max}$ such that for all $t \in \mathbb{N}_{\geq 0}$ and $h^{t+1} \in \mathcal{H}^{t+1}$, $|f_t(h^{t+1})| < f_{max}$.*

*then*

$$\sum_{t=0}^\infty \gamma^t \sum_{i=1}^n \frac{\rho_t^i}{\sum_{j=1}^n \rho_t^j} f_t(H_i^{t+1}) \quad (6)$$

$$\xrightarrow{a.s.} \mathbf{E}\left[\sum_{t=0}^\infty \gamma^t f_t(H^{t+1}) \bigg| H \sim \pi_e\right].$$

*Proof.* Let

$$X_n^t := \sum_{i=1}^n \frac{\rho_t^i}{\sum_{j=1}^n \rho_t^j} \gamma^t f_t(H_i^{t+1}),$$

so that the left side of (6) can be written as $\sum_{t=0}^\infty X_n^t$. First we multiply the numerator and denominator of $X_n^t$ by $\frac{1}{n}$ to get:

$$X_n^t = \frac{\frac{1}{n}\sum_{i=1}^n \gamma^t \rho_t^i f_t(H_i^{t+1})}{\frac{1}{n}\sum_{i=1}^n \rho_t^i}. \quad (7)$$

We will show that the numerator of (7) converges almost surely to the desired value:

$$\frac{1}{n}\sum_{i=1}^n \rho_t^i \gamma^t f_t(H_i^{t+1}) \xrightarrow{a.s.} \mathbf{E}[\gamma^t f_t(H^{t+1})|H^{t+1} \sim \pi_e]. \quad (8)$$

By Lemma 10, which relies on Assumption 1, we have that $\mathbf{E}[\rho_t^i \gamma^t f_t(H_i^{t+1})] = \mathbf{E}[\gamma^t f_t(H^{t+1})|H^{t+1} \sim \pi_e]$. Consider the two cases from the statement of the lemma:

1. **Case 1:** $H_i^{t+1}$ is independent and identically distributed for all $i$, so $\rho_t^i \gamma^t f_t(H_i^{t+1})$ is also independent and identically distributed for all $i$. Therefore by Khintchine's strong law of large numbers, Theorem 6, we have (8).

2. **Case 2:** $H_i^{t+1}$ are *not* necessarily identically distributed since there may be multiple behavior policies, so we cannot directly apply Khintchine's strong law of large numbers. Instead notice that $\rho_t^i$ is bounded by $\beta$ due to Assumption 4, and so $|\rho_t^i \gamma^t f_t(H_i^{t+1})| \leq \beta \gamma^t f_{max}$. So, we can apply Kolmogorov's strong law of large numbers, Corollary 1, to get (8).

Next we show that the denominator of (7) converges almost surely to one:

$$\frac{1}{n}\sum_{i=1}^n \rho_t^i \xrightarrow{a.s.} 1. \quad (9)$$

By Lemma 11, which relies on Assumption 1, we have that $\mathbf{E}[\rho_t^i] = 1$. Again consider the two possible settings:

1. **Case 1:** $H_i^{t+1}$ is independent and identically distributed for all $i$, so $\rho_t^i$ is also independent and identically distributed for all $i$. Therefore by Khintchine's strong law of large numbers we have (9).

2. **Case 2:** Since $\rho_t^i \leq \beta$, we can apply Kolmogorov's strong law of large numbers to get (9).

By applying Property 2 to (8) and (9) we have that for all $t$, $X_n^t \xrightarrow{a.s.} \mathbf{E}\left[\gamma^t f_t(H^{t+1}) \middle| H^{t+1} \sim \pi_e\right]$. So,

1. **Case 1:** Since $X_n^t = 0$ for $t \geq L$ and by Property 3,

$$\sum_{t=0}^{\infty} X_n^t = \sum_{t=0}^{L-1} X_n^t$$
$$\xrightarrow{a.s.} \mathbf{E}\left[\sum_{t=0}^{L-1} \gamma^t f_t(H^{t+1}) \middle| H^{t+1} \sim \pi_e\right]$$
$$= \mathbf{E}\left[\sum_{t=0}^{\infty} \gamma^t f_t(H^{t+1}) \middle| H \sim \pi_e\right].$$

2. **Case 2:** In order to apply Property 4 we must show that there exists a function $g : \mathbb{N}_{\geq 0} \to \mathbb{R}$ such that $\sum_{t=0}^{\infty} g(t) < \infty$ and for all $n \in \mathbb{N}_{>0}$ and $t \in \mathbb{N}_{\geq 0}$, $|X_n^t| \leq g(t)$. The following definition of $g$ satisfies these requirements:

$$g(t) \coloneqq \begin{cases} \gamma^t f_{\max} & \text{if } t < L, \\ 0 & \text{otherwise.} \end{cases}$$

That is,

$$\sum_{t=0}^{\infty} g(t) \leq \begin{cases} \frac{f_{\max}}{1-\gamma} & \text{if } \gamma < 1, \\ L f_{\max} & \text{otherwise,} \end{cases}$$
$$< \infty,$$

since we have assumed that $\gamma$ can only be 1 in the finite-horizon setting, where $L \neq \infty$. Also, $|X_n^t| = 0 = g(t)$ by definition if $t \geq L$ and if $t < L$ then:

$$|X_n^t| \coloneqq \left|\sum_{i=1}^{n} \frac{\rho_t^i}{\sum_{j=1}^{n} \rho_t^j} \gamma^t f_t(H_i^{t+1})\right|$$
$$\leq \gamma^t f_{\max} \sum_{i=1}^{n} \frac{\rho_t^i}{\sum_{j=1}^{n} \rho_t^j}$$
$$= \gamma^t f_{\max}$$
$$= g(t).$$

So, by Property 4, we have (6).

□

Finally, we establish an extension of Lemma 12 that will facilitate its use with sequences that are not quite in the form that it is defined for:

**Lemma 13.** *For all $t \in \mathbb{N}_{\geq 0}$, let $f_t : \mathcal{H}^t \to \mathbb{R}$. If Assumption 1 holds, $f_t = 0$ for all $t \in \mathbb{N}_{\geq L}$, and either:*

- *Case 1: Assumptions 2 and 3 hold.*
  *or*
- *Case 2: Assumption 4 holds and there is a finite $f_{max}$ such that for all $t \in \mathbb{N}_{\geq 0}$ and $h^t \in \mathcal{H}^t$, $|f_t(h^t)| < f_{max}$.*

*then*

$$\sum_{t=0}^{\infty} \gamma^t \sum_{i=1}^{n} \frac{\rho_{t-1}^i}{\sum_{j=1}^{n} \rho_{t-1}^j} f_t(H_i^t) \xrightarrow{a.s.} \mathbf{E}\left[\sum_{t=0}^{\infty} \gamma^t f_t(H^t) \middle| H \sim \pi_e\right]. \quad (10)$$

*Proof.* By removing the first term of the sum and shifting the variable that the sum uses by one, we can rewrite the left side of (10) as

$$\frac{1}{n}\sum_{i=1}^{n} f_0(H_i^0) + \sum_{t=0}^{\infty} \gamma^t \sum_{i=1}^{n} \frac{\rho_t^i}{\sum_{j=1}^{n} \rho_t^j} \gamma f_{t+1}(H_i^{t+1}).$$

We have that

$$\frac{1}{n}\sum_{i=1}^{n} f_0(H_i^0) \xrightarrow{a.s.} \mathbf{E}[f_0(H^0)], \quad (11)$$

by Khintchine's strong law of large numbers in Case 1, and Kolmogorov's strong law of large numbers in Case 2 (since $f_0$ is bounded). Also, by Lemma 12 (where the definition of $f_{t+1}$ in this lemma is used for $f_t$ in our application of Lemma 12) we have that

$$\sum_{t=0}^{\infty} \gamma^t \sum_{i=1}^{n} \frac{\rho_t^i}{\sum_{j=1}^{n} \rho_t^j} \gamma f_{t+1}(H_i^{t+1})$$
$$\xrightarrow{a.s.} \mathbf{E}\left[\sum_{t=0}^{\infty} \gamma^{t+1} f_{t+1}(H^{t+1}) \middle| H \sim \pi_e\right]. \quad (12)$$

So by applying Property 3 to (11) and (12) we have:

$$\frac{1}{n}\sum_{i=1}^{n} f_0(H_i^0) + \sum_{t=0}^{\infty} \gamma^t \sum_{i=1}^{n} \frac{\rho_t^i}{\sum_{j=1}^{n} \rho_t^j} \gamma f_{t+1}(H_i^{t+1})$$
$$\xrightarrow{a.s.} \mathbf{E}[f_0(H^0)] + \mathbf{E}\left[\sum_{t=0}^{\infty} \gamma^{t+1} f_{t+1}(H^{t+1}) \middle| H \sim \pi_e\right]$$
$$= \sum_{t=0}^{0} \mathbf{E}[\gamma^t f_t(H^t)] + \mathbf{E}\left[\sum_{t=1}^{\infty} \gamma^t f_t(H^t) \middle| H \sim \pi_e\right]$$
$$= \mathbf{E}\left[\sum_{t=0}^{\infty} \gamma^t f_t(H^t) \middle| H \sim \pi_e\right].$$

□

## B. Doubly Robust Derivation and Proofs

In this appendix we provide an alternate derivation of the DR estimator using control variates. The idea behind control variates is as follows. Suppose that we would like to estimate $\theta \coloneqq \mathbf{E}[X]$ given a sample of $X$. The obvious estimator would be $\hat{\theta}_1 \coloneqq X$. However, if we have a sample of another random variable, $Y$, with known expected value, $\mathbf{E}[Y]$, then the estimator $\hat{\theta}_2 \coloneqq X - Y + \mathbf{E}[Y]$ may have lower variance. Specifically, while $\text{Var}(\hat{\theta}_1) = \text{Var}(X)$, we have that $\text{Var}(\hat{\theta}_2) = \text{Var}(X) + \text{Var}(Y) - 2\text{Cov}(X, Y)$. So, $\hat{\theta}_2$ has lower variance than $\hat{\theta}_1$ if $2\text{Cov}(X, Y) > \text{Var}(Y)$. Often $Y$ is referred to as the *control variate*. Notice that the optimal control variate is $Y := X$, since then $\text{Var}(\hat{\theta}_2) = 0$. Furthermore, notice that $\hat{\theta}_2$ remains an unbiased estimator of $\theta$ as long as the expected value of $Y$ exists—$\mathbf{E}[\hat{\theta}_2] = \mathbf{E}[X - Y + \mathbf{E}[Y]] = \mathbf{E}[X] - \mathbf{E}[Y] + \mathbf{E}[Y] = \mathbf{E}[X] = \theta$. Control variates have been used before in reinforcement learning to reduce the variance of policy gradient estimates (Bhatnagar et al., 2009), where the control variate was referred to as a *baseline*.

Recall that we have defined the DR estimator in (2) as

$$\text{DR}(D) := \underbrace{\sum_{i=1}^{n} \sum_{t=0}^{\infty} \gamma^t w_t^i R_t^{H_i}}_{X}$$
$$- \underbrace{\sum_{i=1}^{n} \sum_{t=0}^{\infty} \gamma^t \left( w_t^i \hat{q}^{\pi_e} \left( S_t^{H_i}, A_t^{H_i} \right) - w_{t-1}^i \hat{v}^{\pi_e} \left( S_t^{H_i} \right) \right)}_{Y}.$$

In this definition the $X$ term is the *per-decision importance sampling* (PDIS) estimator, which is known to be an unbiased and strongly consistent estimator of $v(\pi_e)$ (Precup et al., 2000; Thomas, 2015b). Also, the control variate, $Y$, is mean zero, i.e., $\mathbf{E}[Y] = 0$. To see why this control variate is reasonable, notice that all of the terms that are multiplied by $\gamma^t w_t^i$ approximately cancel:

$$\hat{q}^{\pi_e} \left( S_t^{H_i}, A_t^{H_i} \right) \approx R_t^{H_i} + \gamma \hat{v}^{\pi_e} \left( S_{t+1}^{H_i} \right).$$

So, $Y$ is a decent approximation of $X$, and therefore $\text{DR}(D)$ will have low variance.

Our derivation of the control variate used by the DR estimator is based on an alternate view of control variates. If we do not know the expected value of the control variate, $Y$, but we have another random variable, $Z$, such that $\mathbf{E}[Z] = \mathbf{E}[Y]$, then we can use the unbiased estimator $\hat{\theta}_3 = X - Y + Z$. The variance of this estimator is given by $\text{Var}(\hat{\theta}_3) = \text{Var}(X) + \text{Var}(Y - Z) - 2\text{Cov}(X, Y - Z)$. So, if $Y \approx X$ and $Z$ has low variance, then this estimator may have lower variance than $\hat{\theta}_1$. Technically, this is an ordinary application of control variates using $Y - Z$ as the mean-zero control variate. We derive DR using this alternate view.

We begin with the *per-decision importance sampling* (PDIS) estimator, which is known to be an unbiased and strongly consistent estimator of $v(\pi_e)$ (Precup et al., 2000;

Thomas, 2015b). The PDIS estimator is given by:

$$\text{PDIS}(D) := \frac{1}{n} \sum_{i=1}^{n} \sum_{t=0}^{\infty} \rho_t^i \gamma^t R_t^{H_i}.$$

In order to reduce the variance of this estimator we will subtract a control variate that we expect to be highly correlated with the PDIS estimator, and then add back in the expected value of the control variate:

$$\underbrace{\frac{1}{n} \sum_{i=1}^{n} \sum_{t=0}^{\infty} \rho_t^i \gamma^t R_t^{H_i}}_{\text{PDIS estimator, }X} - \underbrace{\frac{1}{n} \sum_{i=1}^{n} \sum_{t=0}^{\infty} \rho_t^i \gamma^t \hat{r}^{\pi_e}(S_t^{H_i}, A_t^{H_i}, 0)}_{\text{control variate, }Y}$$
$$+ \underbrace{\mathbf{E}\left[ \frac{1}{n} \sum_{i=1}^{n} \sum_{t=0}^{\infty} \rho_t^i \gamma^t \hat{r}^{\pi_e}(S_t^{H_i}, A_t^{H_i}, 0) \right]}_{\mathbf{E}[\text{control variate}] = \mathbf{E}[Y]}. \quad (13)$$

Here we expect the control variate to be similar to the PDIS estimator if the model's reward predictions are accurate, i.e., if $R_t^{H_i} \approx \hat{r}^{\pi_e}(S_t^{H_i}, A_t^{H_i}, 0)$.

If it could be used, (13) would be an extremely low-variance estimator of $v(\pi_e)$ since $X - Y$ would usually be near-zero and $\mathbf{E}[Y]$ is a constant that is near $v(\pi_e)$. However, $\mathbf{E}[\text{control variate}]$ is not known, and so we cannot use (13) directly. Although estimating $\mathbf{E}[Y]$ is nearly as hard as estimating $v(\pi_e)$, it is marginally easier. It is easier because $v(\pi_e)$ uses the unknown transition and reward functions of the MDP to produce the distribution of rewards at each time step, while $\mathbf{E}[Y]$ uses the known approximate model's transition and reward function for the last transition before each reward occurs. We can therefore estimate $\mathbf{E}[Y]$ using an unbiased estimator that typically has lower variance than the control variate. In the alternate view of control variates this new term will be $Z$:

$$\underbrace{\frac{1}{n} \sum_{i=1}^{n} \sum_{t=0}^{\infty} \rho_t^i \gamma^t R_t^{H_i}}_{\text{PDIS estimator, }X} - \underbrace{\frac{1}{n} \sum_{i=1}^{n} \sum_{t=0}^{\infty} \rho_t^i \gamma^t \hat{r}^{\pi_e} \left( S_t^{H_i}, A_t^{H_i}, 0 \right)}_{\text{control variate, }Y}$$
$$+ \underbrace{\frac{1}{n} \sum_{i=1}^{n} \sum_{t=0}^{\infty} \rho_{t-1}^i \gamma^t \hat{r}^{\pi_e} \left( S_t^{H_i}, 0 \right)}_{Z}. \quad (14)$$

Here we expect the $Z$ term to have lower variance than the $Y$ term because for each $i$ and $t$ it only depends on actions $A_1^{H_i}, \ldots, A_{t-1}^{H_i}$ and not $A_t^{H_i}$. This is reflected in its use of $\rho_{t-1}^i$ rather than $\rho_t^i$. Before continuing our derivation we verify that $\mathbf{E}[Y] = \mathbf{E}[Z]$ if Assumption 1 holds:

$$\begin{aligned}\mathbf{E}[Z] =& \mathbf{E}\left[\frac{1}{n}\sum_{i=1}^{n}\sum_{t=0}^{\infty}\rho_{t-1}^{i}\gamma^{t}\hat{r}^{\pi_{e}}\left(S_{t}^{H_{i}}, 0\right)\right]\\ \stackrel{(a)}{=}& \mathbf{E}\left[\frac{1}{n}\sum_{i=1}^{n}\sum_{t=0}^{\infty}\rho_{t-1}^{i}\gamma^{t}\frac{\pi_{e}\left(A_{t}^{H_{i}}|S_{t}^{H_{i}}\right)}{\pi_{i}\left(A_{t}^{H_{i}}|S_{t}^{H_{i}}\right)}\hat{r}^{\pi_{e}}\left(S_{t}^{H_{i}}, A_{t}^{H_{i}}, 0\right)\right]\\ =& \mathbf{E}\left[\frac{1}{n}\sum_{i=1}^{n}\sum_{t=0}^{\infty}\rho_{t}^{i}\gamma^{t}\hat{r}^{\pi_{e}}\left(S_{t}^{H_{i}}, A_{t}^{H_{i}}, 0\right)\right]\\ =& \mathbf{E}[Y],\end{aligned}$$

where **(a)** comes from Lemma 6.

So far, in (14), we have introduced a control variate into PDIS that we expect might reduce the variance of the estimator a little without introducing bias. However, it will still have high variance because $Z$ is a high-variance estimator of $\mathbf{E}[Y]$. To overcome this, we can introduce another control variate into $Z$ to make it a lower-variance estimator of $\mathbf{E}[Y]$. So, we introduce another control variate:

$$\underbrace{\frac{1}{n}\sum_{i=1}^{n}\sum_{t=0}^{\infty}\rho_{t}^{i}\gamma^{t}R_{t}^{H_{i}}}_{X} - \underbrace{\frac{1}{n}\sum_{i=1}^{n}\sum_{t=0}^{\infty}\rho_{t}^{i}\gamma^{t}\hat{r}^{\pi_{e}}\left(S_{t}^{H_{i}}, A_{t}^{H_{i}}, 0\right)}_{Y}$$

$$+ \underbrace{\frac{1}{n}\sum_{i=1}^{n}\sum_{t=0}^{\infty}\rho_{t-1}^{i}\gamma^{t}\hat{r}^{\pi_{e}}\left(S_{t}^{H_{i}}, 0\right)}_{Z}$$

$$- \underbrace{\frac{1}{n}\sum_{i=1}^{n}\sum_{t=0}^{\infty}\rho_{t-1}^{i}\gamma^{t}\hat{r}^{\pi_{e}}\left(S_{t-1}^{H_{i}}, A_{t-1}^{H_{i}}, 1\right)}_{\text{new control variate},Y'}$$

$$+ \underbrace{\frac{1}{n}\sum_{i=1}^{n}\sum_{t=0}^{\infty}\rho_{t-2}^{i}\gamma^{t}\hat{r}^{\pi_{e}}\left(S_{t-1}^{H_{i}}, 1\right)}_{Z'}.$$

Here $\mathbf{E}[Z'] = \mathbf{E}[Y']$ (although we omit to proof of this claim), $Y'$ is similar to $Z$ and so it serves as a good control variate therefor, and $Z'$ will usually have lower variance than $Y'$ because it uses $\rho_{t-2}^{i}$ rather than $\rho_{t-1}^{i}$. However, now $Z'$ is a high-variance estimator of $\mathbf{E}[Y']$. We therefore introduce a control variate for $Z'$, and this process repeats. This process of introducing control variates eventually terminates when the new control variate is not random. The resulting estimator is (we call this estimator $\text{DR}(D)$ because we will show that it is equivalent to (2)):

$$\begin{aligned}\text{DR}(D) =& \frac{1}{n}\sum_{i=1}^{n}\sum_{t=0}^{\infty}\rho_{t}^{i}\gamma^{t}R_{t}^{H_{i}} \quad (15)\\ & - \frac{1}{n}\sum_{i=1}^{n}\sum_{t=0}^{\infty}\gamma^{t}\sum_{\tau=0}^{t}\rho_{\tau}^{i}\hat{r}^{\pi_{e}}\left(S_{\tau}^{H_{i}}, A_{\tau}^{H_{i}}, t-\tau\right)\\ & + \frac{1}{n}\sum_{i=1}^{n}\sum_{t=0}^{\infty}\gamma^{t}\sum_{\tau=0}^{t}\rho_{\tau-1}^{i}\hat{r}^{\pi_{e}}\left(S_{\tau}^{H_{i}}, t-\tau\right).\end{aligned}$$

Next we will combine the $\hat{r}$ terms into $\hat{v}$ and $\hat{q}$ terms to get a more succinct expression. To this end, we will use the property that $\sum_{i=0}^{\infty}\sum_{j=0}^{i}f(i,j) = \sum_{j=0}^{\infty}\sum_{i=j}^{\infty}f(i,j)$ to change the order of the sums over $t$ and $\tau$. We also split $\gamma^{t}$ into $\gamma^{\tau}\gamma^{t-\tau}$:

$$\begin{aligned}\text{DR}(D) =& \frac{1}{n}\sum_{i=1}^{n}\sum_{t=0}^{\infty}\rho_{t}^{i}\gamma^{t}R_{t}^{H_{i}}\\ & - \frac{1}{n}\sum_{i=1}^{n}\sum_{\tau=0}^{\infty}\rho_{\tau}^{i}\gamma^{\tau}\sum_{t=\tau}^{\infty}\gamma^{t-\tau}\hat{r}^{\pi_{e}}\left(S_{\tau}^{H_{i}}, A_{\tau}^{H_{i}}, t-\tau\right)\\ & + \frac{1}{n}\sum_{i=1}^{n}\sum_{\tau=0}^{\infty}\rho_{\tau-1}^{i}\gamma^{\tau}\sum_{t=\tau}^{\infty}\gamma^{t-\tau}\hat{r}^{\pi_{e}}\left(S_{\tau}^{H_{i}}, t-\tau\right).\end{aligned}$$

Next we perform a change of variable using $j = t - \tau$ to replace $t$:

$$\begin{aligned}\text{DR}(D) =& \frac{1}{n}\sum_{i=1}^{n}\sum_{t=0}^{\infty}\rho_{t}^{i}\gamma^{t}R_{t}^{H_{i}}\\ & - \frac{1}{n}\sum_{i=1}^{n}\sum_{\tau=0}^{\infty}\rho_{\tau}^{i}\gamma^{\tau}\sum_{j=0}^{\infty}\gamma^{j}\hat{r}^{\pi_{e}}\left(S_{\tau}^{H_{i}}, A_{\tau}^{H_{i}}, j\right)\\ & + \frac{1}{n}\sum_{i=1}^{n}\sum_{\tau=0}^{\infty}\rho_{\tau-1}^{i}\gamma^{\tau}\sum_{j=0}^{\infty}\gamma^{j}\hat{r}^{\pi_{e}}\left(S_{\tau}^{H_{i}}, j\right)\\ =& \frac{1}{n}\sum_{i=1}^{n}\sum_{t=0}^{\infty}\rho_{t}^{i}\gamma^{t}R_{t}^{H_{i}}\\ & - \frac{1}{n}\sum_{i=1}^{n}\sum_{\tau=0}^{\infty}\rho_{\tau}^{i}\gamma^{\tau}\hat{q}^{\pi_{e}}\left(S_{\tau}^{H_{i}}, A_{\tau}^{H_{i}}\right)\\ & + \frac{1}{n}\sum_{i=1}^{n}\sum_{\tau=0}^{\infty}\rho_{\tau-1}^{i}\gamma^{\tau}\hat{v}^{\pi_{e}}\left(S_{\tau}^{H_{i}}\right).\end{aligned}$$

Replacing the variable $\tau$ with $t$ and using $w_{t}^{i} = \frac{\rho_{t}^{i}}{n}$ we get

that:

$$\mathrm{DR}(D) = \sum_{i=1}^{n} \sum_{t=0}^{\infty} \gamma^t w_t^i R_t^{H_i}$$
$$- \sum_{i=1}^{n} \sum_{t=0}^{\infty} \gamma^t \left( w_t^i \hat{q}^{\pi_e} \left( S_t^{H_i}, A_t^{H_i} \right) - w_{t-1}^i \hat{v}^{\pi_e} \left( S_t^{H_i} \right) \right),$$

which is (2).

The original derivation of the DR estimator (Jiang & Li, 2015) required the horizon to be finite and known. Our derivation makes neither of these assumptions. That is, it allows for infinite or indefinite horizons and for finite horizons where the horizon is not known. If the horizon, $L$, is finite and known, then one should ensure that the model uses all of the available information, including the known horizon and time step. In the next section we show that if $L$ is finite and known, then our non-recursive definition of the DR estimator is equivalent to the recursive form of (Jiang & Li, 2015).

### B.1. Equivalence of DR Definitions

In this section we show that our non-recursive definition of the DR estimator is equivalent to the recursive definition provided by Jiang & Li (2015) when the horizon is finite and known.

**Theorem 8.** *(2) is equivalent to the DR estimator presented by Jiang & Li (2015) if the finite horizon, L, of the MDP is known.*

*Proof.* Jiang & Li (2015) define the DR estimator for a single trajectory (i.e., $n = 1$) as the last element, $X_L$, of a sequence, $(X_i)_{i=0}^{L}$. This sequence is defined by the following recurrence relation. Let $X_0 \coloneqq 0$ and for all $k \in \{1, \ldots, L\}$ let

$$X_k \coloneqq \hat{v}^{\pi_e}(S_{L-k}) + \frac{\pi_e(A_{L-k}|S_{L-k})}{\pi_1(A_{L-k}|S_{L-k})} \bigg( R_{L-k} + \gamma X_{k-1}$$
$$- \hat{q}^{\pi_e}(S_{L-k}, A_{L-k}) \bigg).$$

As in the definition of $\mathrm{DR}(D)$ in (2), Jiang & Li (2015) define the DR estimator for multiple trajectories to be the average of the estimator for each trajectory individually. So, to show that their recursive definition and our definition are equivalent, we need only show that they are equivalent when there is a single trajectory.

Since hereafter in this proof we deal with only a single trajectory, we drop the superscripts that we use to specify the trajectory, i.e., we write $\rho_t$ rather than $\rho_t^1$. Also let $\pi_b \coloneqq \pi_1$ denote the single behavior policy. For further brevity, let

$$\pi_b^e(t) \coloneqq \frac{\pi_e(A_t|S_t)}{\pi_b(A_t|S_t)}.$$

First, notice that we can rewrite (2) for the single-trajectory finite-horizon setting as:

$$\mathrm{DR}(D) = \sum_{t=0}^{L-1} \gamma^t \rho_t R_t - \sum_{t=0}^{L-1} \gamma^t \rho_t \hat{q}^{\pi_e}(S_t, A_t)$$
$$+ \sum_{t=0}^{L-1} \gamma^t \rho_{t-1} \hat{v}^{\pi_e}(S_t)), \quad (16)$$

since $S_L$ is surely the absorbing state and so $R_t$, $\hat{q}^{\pi_e}(S_t, A_t)$, and $\hat{v}^{\pi_e}(S_t)$ are all zero for $t \geq L$. To verify that this definition is equivalent to $X_L$, we will define another sequence, $(Y_i)_{i=1}^{L}$, such that $X_i = Y_i$ for all $i \in \{1, \ldots, L\}$ and such that $Y_L = \mathrm{DR}(D)$ trivially.

Let

$$Y_k \coloneqq \frac{\sum_{t=L-k}^{L-1} \gamma^t \left[ \rho_t \left( R_t - \hat{q}^{\pi_e}(S_t, A_t) \right) + \rho_{t-1} \hat{v}^{\pi_e}(S_t) \right]}{\gamma^{L-k} \rho_{L-k-1}}.$$

Notice that $Y_L$ is identical to (16) since $\gamma^{L-L} \rho_{L-L-1} = 1$. So, all that remains is to show that $Y_k = X_k$ for all $k \in \{1, \ldots, L\}$. We will show this using a proof by induction.

For the base case, $k = 1$, it is straightforward to verify that $X_1 = Y_1$. For the inductive step we assume the inductive hypothesis that $X_{k-1} = Y_{k-1}$ and show that then $X_k = Y_k$:

$$X_k \coloneqq \hat{v}^{\pi_e}(S_{L-k}) + \pi_b^e(L-k) \bigg( R_{L-k} + \gamma X_{k-1}$$
$$- \hat{q}^{\pi_e}(S_{L-k}, A_{L-k}) \bigg)$$
$$= \hat{v}^{\pi_e}(S_{L-k}) + \pi_b^e(L-k) \bigg( R_{L-k} + \gamma Y_{k-1}$$
$$- \hat{q}^{\pi_e}(S_{L-k}, A_{L-k}) \bigg).$$

Substituting in the definition of $Y_{k-1}$ and performing algebraic manipulations we have that:

$$X_k = \hat{v}^{\pi_e}(S_{L-k}) + \pi_b^e(L-k) R_{L-k} + \frac{\pi_b^e(L-k)}{\gamma^{L-k} \rho_{L-k}}$$
$$\times \sum_{t=L-k+1}^{L-1} \gamma^t \left[ \rho_t (R_t - \hat{q}^{\pi_e}(S_t, A_t)) + \rho_{t-1} \hat{v}^{\pi_e}(S_t) \right]$$
$$- \pi_b^e(L-k) \hat{q}^{\pi_e}(S_{L-k}, A_{L-k}),$$

where $\times$ denotes that a line was split into multiple lines (we do not use cross-products anywhere in this paper). Since

$$\frac{\pi_b^e(L-k)}{\rho_{L-k}} = \frac{1}{\rho_{L-k-1}},$$

and by reordering terms, we have that

$$X_k = \pi_b^e(L-k)(R_{L-k} - \hat{q}^{\pi_e}(S_{L-k}, A_{L-k})) + \hat{v}^{\pi_e}(S_{L-k})$$

$$+ \frac{\sum_{t=L-k+1}^{L-1} \gamma^t \left[ \rho_t (R_t - \hat{q}^{\pi_e}(S_t, A_t)) + \rho_{t-1} \hat{v}^{\pi_e}(S_t) \right]}{\gamma^{L-k} \rho_{L-k-1}}.$$

Adding one more element to the summation so that it starts at $t = L - k$, and then explicitly subtracting off this additional term we have that:

$$X_k = \pi_b^e(L-k)(R_{L-k} - \hat{q}^{\pi_e}(S_{L-k}, A_{L-k})) + \hat{v}^{\pi_e}(S_{L-k})$$

$$+ \frac{\sum_{t=L-k}^{L-1} \gamma^t \left[ \rho_t (R_t - \hat{q}^{\pi_e}(S_t, A_t)) + \rho_{t-1} \hat{v}^{\pi_e}(S_t) \right]}{\gamma^{L-k} \rho_{L-k-1}}$$

$$- \frac{\gamma^{L-k}}{\gamma^{L-k} \rho_{L-k-1}} \Bigg[ \rho_{L-k} (R_{L-k} - \hat{q}^{\pi_e}(S_{L-k}, A_{L-k}))$$

$$+ \rho_{L-k-1} \hat{v}^{\pi_e}(S_{L-k}) \Bigg].$$

Canceling several $\gamma$ and $\rho$ terms, we have that:

$$X_k = \frac{\sum_{t=L-k}^{L-1} \gamma^t \left[ \rho_t (R_t - \hat{q}^{\pi_e}(S_t, A_t)) + \rho_{t-1} \hat{v}^{\pi_e}(S_t) \right]}{\gamma^{L-k} \rho_{L-k-1}}$$

$$= Y_k.$$

$\square$

### B.2. DR is Unbiased

While Jiang & Li (2015) showed that the DR estimator (with finite horizon) is an unbiased estimator of $v(\pi_e)$, in this section we show that the DR estimator (without assumptions about the horizon) is an unbiased estimator of $v(\pi_e)$.

**Theorem 9** (DR – unbiased estimator). *If Assumption* 1 *holds, then* $\mathbf{E}[\mathrm{DR}(D)] = v(\pi_e)$.

*Proof.* This result was shown previously for the known finite horizon setting (Jiang & Li, 2015), but has not been shown before for the other settings. Because we will use some steps of this proof in later proofs, the majority of this proof is relegated to a lemma.

$$\mathbf{E}[\mathrm{DR}(D)] = \mathbf{E}\left[ \frac{1}{n} \sum_{i=1}^n \mathrm{DR}_i(D) \right]$$

$$\stackrel{(a)}{=} \frac{1}{n} \sum_{i=1}^n v(\pi_e)$$

$$= v(\pi_e),$$

where **(a)** comes from Lemma 7. $\square$

### B.3. Conditions for Consistency of DR

In this section we show that the DR estimator is a strongly consistent estimator of $v(\pi_e)$ given mild technical assumptions and that there is only one behavior policy (Theorem 10) or that the importance weights are bounded (Theorem 11).

**Theorem 10** (DR – strongly consistent estimator for one behavior policy). *If Assumptions* 1 *and* 2 *hold then* $\mathrm{DR}(D) \xrightarrow{a.s.} v(\pi_e)$.

*Proof.* This proof is a relatively straightforward application of the law of large numbers.

We have from Lemma 7 that $\mathbf{E}[\mathrm{DR}_i(D)] = v(\pi_e)$ for all $i \in \{1, \ldots, n\}$. By Assumption 2, $\{\mathrm{DR}_i(D)\}_{i=1}^n$ is a set of $n$ independent *and identically distributed* random variables (since $H_i \sim \pi_1$ for all $i$, and $\mathrm{DR}_i(D)$ only depends on $H_i$). We can therefore conclude by Khintchine's strong law of large numbers, Theorem 6, that $\mathrm{DR}(D) \xrightarrow{a.s.} v(\pi_e)$. $\square$

**Theorem 11** (DR – strongly consistent estimator for many behavior policies). *If Assumptions* 1 *and* 4 *hold then* $\mathrm{DR}(D) \xrightarrow{a.s.} v(\pi_e)$.

*Proof.* We have from Lemma 7 that $\mathbf{E}[\mathrm{DR}_i(D)] = v(\pi_e)$ for all $i \in \{1, \ldots, n\}$. However, $\{\mathrm{DR}_i(D)\}_{i=1}^n$ is a set of $n$ independent but not necessarily identically distributed random variables, so we cannot apply Khintchine's strong law of large numbers. Instead, we will apply Kolmogorov's strong law of large numbers, which requires each random variable, $\mathrm{DR}_i(D)$, to be bounded.

We have that:

$$\mathrm{DR}_i(D) = \sum_{t=0}^\infty \gamma^t \rho_t^i R_t^{H_i} - \sum_{t=0}^\infty \gamma^t \rho_t^i \hat{q}^{\pi_e}\left(S_t^{H_i}, A_t^{H_i}\right)$$

$$+ \sum_{t=0}^\infty \gamma^t \rho_{t-1}^i \hat{v}^{\pi_e}\left(S_t^{H_i}\right)$$

$$= \sum_{t=0}^\infty \gamma^t \rho_t^i R_t^{H_i}$$

$$- \sum_{t=0}^\infty \gamma^t \rho_t^i \underbrace{\sum_{\tau=0}^\infty \gamma^\tau \hat{r}^{\pi_e}\left(S_t^{H_i}, A_t^{H_i}, \tau\right)}_{=:\hat{q}^{\pi_e}\left(S_t^{H_i}, A_t^{H_i}\right)}$$

$$+ \sum_{t=0}^\infty \gamma^t \rho_{t-1}^i \underbrace{\sum_{\tau=0}^\infty \gamma^\tau \hat{r}^{\pi_e}\left(S_t^{H_i}, \tau\right)}_{=:\hat{v}^{\pi_e}\left(S_t^{H_i}\right)}.$$

So,

$$|\mathrm{DR}_i(D)| \leq 3\beta r_{\max}^\star \sum_{t=0}^{L} \gamma^t \sum_{\tau=0}^{L} \gamma^\tau$$
$$< \infty,$$

since either $L < \infty$ or $\gamma \in [0, 1)$. So, $\mathrm{DR}_i(D)$ is bounded above and below and thus we can apply Kolmogorov's strong law of large numbers (Corollary 1) to conclude that $\mathrm{DR}(D) \xrightarrow{\text{a.s.}} v(\pi_e)$. $\square$

## C. Weighted Doubly Robust Proofs

### C.1. Proof of Theorem 1

In this section we prove Theorem 1, which states that $\mathrm{WDR}(D)$ is a strongly consistent estimator of $v(\pi_e)$ if Assumptions 1, 2, and 3 hold.

First, notice that we can rewrite the WDR estimator as:

$$\mathrm{WDR}(D) := \underbrace{\sum_{t=0}^{\infty} \gamma^t \sum_{i=1}^{n} \frac{\rho_t^i}{\sum_{j=1}^{n} \rho_t^j} R_t^{H_i}}_{=: \mathrm{CWPDIS}(D)} \quad (17)$$
$$- \underbrace{\sum_{t=0}^{\infty} \gamma^t \sum_{i=1}^{n} \frac{\rho_t^i}{\sum_{j=1}^{n} \rho_t^j} \hat{q}^{\pi_e}\left(S_t^{H_i}, A_t^{H_i}\right)}_{=: X_n}$$
$$+ \underbrace{\sum_{t=0}^{\infty} \gamma^t \sum_{i=1}^{n} \frac{\rho_{t-1}^i}{\sum_{j=1}^{n} \rho_{t-1}^j} \hat{v}^{\pi_e}\left(S_t^{H_i}\right)}_{=: Y_n}.$$

We have from Lemma 12 that

$$\mathrm{CWPDIS}(D) \xrightarrow{\text{a.s.}} \mathbf{E}\left[\sum_{t=0}^{\infty} \gamma^t R_t^H | H \sim \pi_e\right]$$
$$= v(\pi_e), \quad (18)$$

which has been shown before (Thomas, 2015b, Theorem 13). Also by Lemma 12 we have that

$$X_n \xrightarrow{\text{a.s.}} \mathbf{E}\Big[\sum_{t=0}^{\infty} \gamma^t \hat{q}^{\pi_e}\left(S_t^H, A_t^H\right) \Big| H \sim \pi_e\Big], \quad (19)$$

and by Lemma 13 we have that

$$Y_n \xrightarrow{\text{a.s.}} \mathbf{E}\Big[\sum_{t=0}^{\infty} \gamma^t \hat{v}^{\pi_e}\left(S_t^H\right) \Big| H \sim \pi_e\Big]$$
$$= \mathbf{E}\Big[\sum_{t=0}^{\infty} \gamma^t \underbrace{\sum_{j=0}^{\infty} \gamma^j \hat{r}^{\pi_e}\left(S_t^H, j\right)}_{= \hat{v}^{\pi_e}(S_t^H)} \Big| H \sim \pi_e\Big]$$
$$= \mathbf{E}\Big[\sum_{t=0}^{\infty} \gamma^t \sum_{j=0}^{\infty} \gamma^j \underbrace{\sum_{a \in \mathcal{A}} \pi_e\left(a | S_t^H\right) \hat{r}^{\pi_e}\left(S_t^H, a, j\right)}_{= \hat{r}^{\pi_e}(S_t^H, j)} \Big| H \sim \pi_e\Big]$$
$$= \mathbf{E}\Big[\sum_{t=0}^{\infty} \gamma^t \underbrace{\sum_{j=0}^{\infty} \gamma^j \hat{r}^{\pi_e}\left(S_t^H, A_t^H, j\right)}_{= \hat{q}^{\pi_e}(S_t^H, A_t^H)} \Big| H \sim \pi_e\Big]$$
$$= \mathbf{E}\Big[\sum_{t=0}^{\infty} \gamma^t \hat{q}^{\pi_e}\left(S_t^H, A_t^H\right) \Big| H \sim \pi_e\Big]. \quad (20)$$

So, by applying Property 3 to (18), (19), and (20) we have that $\mathrm{WDR}(D) \xrightarrow{\text{a.s.}} v(\pi_e)$.

### C.2. Proof of Theorem 2

In this section we prove Theorem 2, which states that if Assumptions 1 and 4 hold then

$$\mathrm{WDR}(D) \xrightarrow{\text{a.s.}} v(\pi_e).$$

Recall that WDR can be defined as in (17). First we apply Lemma 12 to the $\mathrm{CWPDIS}(D)$ term, which uses $f_t(H_i^{t+1}) = R_t^{H_i}$, which is bounded since $|R_t^{H_i}| \leq r_{\max}^\star$. The result of Lemma 12 is that

$$\mathrm{CWPDIS}(D) \xrightarrow{\text{a.s.}} \mathbf{E}\left[\sum_{t=0}^{\infty} \gamma^t R_t^H | H \sim \pi_e\right]$$
$$= v(\pi_e). \quad (21)$$

Next we apply Lemma 12 to the $X_n$ term, which uses $f_t(H_i^{t+1}) = \hat{q}^{\pi_e}\left(S_t^{H_i}, A_t^{H_i}\right)$, which is bounded since

$$\left|\hat{q}^{\pi_e}\left(S_t^{H_i}, A_t^{H_i}\right)\right| \leq \begin{cases} \frac{r_{\max}^\star}{1-\gamma} & \text{if } L = \infty \\ L r_{\max} & \text{otherwise.} \end{cases}$$

The result of applying Lemma 12 to $X_n$ is that

$$X_n \xrightarrow{\text{a.s.}} \mathbf{E}\Big[\sum_{t=0}^{\infty} \gamma^t \hat{q}^{\pi_e}\left(S_t^H, A_t^H\right) \Big| H \sim \pi_e\Big]. \quad (22)$$

Lastly, we apply Lemma 13 to the $Y_n$ term, which uses $f_t(H_i^t) = \hat{v}^{\pi_e}\left(S_t^{H_i}\right)$, which is bounded since

$$\left|\hat{v}^{\pi_e}\left(S_t^{H_i}\right)\right| \leq \begin{cases} \frac{r_{\max}^\star}{(1-\gamma)} & \text{if } L = \infty \\ L r_{\max}^\star & \text{otherwise.} \end{cases}$$

The result of applying Lemma 13 to $Y_n$ is that

$$Y_n \xrightarrow{\text{a.s.}} \mathbf{E}\Big[\sum_{t=0}^{\infty} \gamma^t \hat{v}^{\pi_e}\left(S_t^H\right) \Big| H \sim \pi_e\Big]$$

$$\stackrel{(a)}{=} \mathbf{E}\Big[\sum_{t=0}^{\infty} \gamma^t \hat{q}^{\pi_e}\left(S_t^H, A_t^H\right) \Big| H \sim \pi_e\Big], \quad (23)$$

where **(a)** comes from the same derivation that was used in (20). So, by applying Property 3 to (21), (22), and (23) we have that $\text{WDR}(D) \xrightarrow{\text{a.s.}} v(\pi_e)$.

## D. Extended Empirical Studies (WDR)

In this section we provide a detailed description of our experiments comparing the WDR estimator to various importance sampling estimators (IS, PDIS, WIS, CWPDIS), as well as DR and AM. We performed experiments using three domains: ModelFail, ModelWin, and a gridworld. We will describe each domain, then describe the experimental setup, and then present empirical results. All three domains have a finite horizon and use $\gamma = 1.0$.

### D.1. The ModelFail Domain

The ModelFail domain was constructed so that the model would fail to converge to the true MDP. One way that this can happen is if the model uses function approximation, so that it cannot represent the true MDP. Another way that this can happen is if there is some partial observability, which is common in real applications. We therefore construct a domain where the true underlying MDP has three states (plus the terminal absorbing state), but where the agent cannot tell the difference between any of the states.

The MDP used by ModelFail is depicted in Figure 3. Although the MDP has three states (denoted by circles) plus the terminal absorbing state (denoted by the double-circle), the agent does not observe which state it is in—it only sees a single state. The agent begins in the left-most state, where it has two actions available. The first action always takes it to the upper state, while the second always takes in to the lower state. In both cases, the agent receives no reward.

At time $t = 1$, the agent is always in the upper or lower state (although it cannot tell the difference between them and the initial state), and it must select between two possible actions. Both actions always have the same effect—the agent transitions to the terminal absorbing state. However, if the agent was in the upper state, $R_1 = 1$, while $R_1 = -1$ if the agent was in the lower state. The horizon is $L = 2$ since $S_2 = \overset{\infty}{s}$ always.

The behavior policy selects $a_1$ with probability approximately $0.88$ and $a_2$ with probability approximately $0.12$ (these probabilities were chosen arbitrarily by using weights of $1$ and $-1$ with softmax action selection, and were not optimized). The evaluation policy does the opposite—it selects $a_1$ with probability approximately $0.12$ and $a_2$ with probability approximately $0.88$.

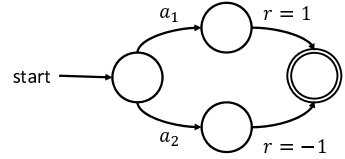

Figure 3: ModelFail MDP.

Consider what happens when we try to model this MDP based on the observations produced by running the behavior policy to produce an infinite number of trajectories (without trying to infer anything about the true underlying structure of the MDP). Recall that we observe only a single state. First consider the transition dynamics: half of the time either action causes a transition back to the single state, while half of the time the agent transitions to the absorbing state. Next consider the rewards: half of the time the agent receives no reward, with probability $0.88/2$ it receives a reward of $1$, and with probability $0.12/2$ it receives a reward of $-1$, and these rewards appear completely uncorrelated with the action that was selected (since non-zero rewards occur at time $t = 1$ and $A_1$ has no bearing on rewards or state transitions). So, from the model's point of view, the actions have no impact on state transitions or rewards, and so every policy is equally good and will produce an expected return of $0.38$, while in reality an optimal policy will produce an expected return of $0.5$ and a pessimal policy will produce an expected return of $-0.5$.

We provided the model with the true horizon, $L = 2$, so that its predictions of $R_t$ are zero for $t \geq 2$.

### D.2. The ModelWin Domain

This domain was constructed so that the approximate model of the MDP would quickly converge to the true MDP, while importance sampling based approaches like DR and WDR would continue to have high variance. Recall from our discussion in Section 6 that DR and WDR will be equal to a simple model-based approach if the approximate MDP is perfect and state transition and rewards are deterministic. To avoid this, the ModelWin domain has stochastic state transitions that cause the **(b)** term in (3) to not necessarily be zero.

The ModelWin MDP is depicted in Figure 4. Unlike the ModelFail domain, the agent observes the true underlying states of the ModelWin MDP, of which there are three, plus a terminal absorbing state (not pictured). The agent always begins in $s_1$, where it must select between two actions. The first action, $a_1$, causes the agent to transition to $s_2$ with probability $0.4$ and $s_3$ with probability $0.6$. The second action, $a_2$, does the opposite: the agent transitions to $s_2$ with probability $0.6$ and $s_3$ with probability $0.4$. If the agent transitions to $s_2$, then it receives a reward of $1$, and if it transitions to $s_3$ it receives a reward of $-1$. In states $s_2$ and $s_3$, the agent has two possible actions, but both always produce a reward of zero and a deterministic transition back to $s_1$. The horizon is set to $L = 20$, so, $S_{20} = \overset{\infty}{s}$ always.[12]

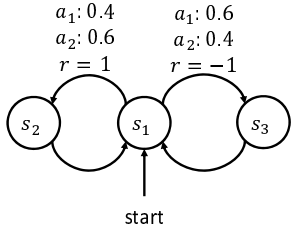

Figure 4: ModelWin MDP.

To see why DR and WDR struggle on this domain, consider what happens if the approximate model is perfect and the agent takes action $a_1$ in state $s_1$. In our discussion of (3) we concluded that DR and WDR will perform well if $R_1 = q^{\pi_e}(s_1, a_1) - \gamma \hat{v}^{\pi_e}(S')$, where $S'$ is the state that the agent transitions to after taking action $a_1$ in state $s_1$, which is a random variable. Consider the two values that the right side can take, depending on whether $S' = s_2$ or $S' = s_3$. It can be either $\hat{q}^{\pi_e}(s_1, a_1) - \gamma \hat{v}^{\pi_e}(s_2)$ or $\hat{q}^{\pi_e}(s_1, a_1) - \gamma \hat{v}^{\pi_e}(s_3)$. Since $\hat{v}^{\pi_e}(s_2) = \hat{v}^{\pi_e}(s_3)$, these two statements are equal—the prediction of $R_1$ will be the same regardless of whether the agent transitions to $s_2$ or $s_3$, and so its prediction must sometimes be wrong (since the rewards differ depending on whether the agent transitions to $s_2$ or $s_3$). So, term **(b)** in (3) will not be zero—the control variate used by DR and WDR does not perfectly cancel with the PDIS (or CWPDIS) term. If $w_t^i$ is large, then this will produce high variance. In order to make $w_t^i$ large, we need only make the horizon long and the behavior and evaluation policies dissimilar.

The behavior and evaluation policies both select actions uniformly randomly in states $s_2$ and $s_3$. However, in $s_1$ the behavior policy takes action $a_1$ with probability approximately $0.73$ and action $a_2$ with probability approximately $0.27$, while the evaluation policy does the opposite—it takes action $a_1$ with probability approximately $0.27$ and action $a_2$ with probability approximately $0.73$ (these probabilities come from using softmax action selection with weights of 1 and 0).

---

[12] Technically, implementing the horizon of $L = 20$ requires the states to be augmented to include the current time step so that state transitions are Markovian. The approximate model is provided with the time step and the horizon.

As in the ModelFail domain, for the ModelWin domain we provided the approximate model with the true horizon of the MDP, $L = 20$, so that its predictions of $R_t$ were zero for $t \geq 20$.

### D.3. The Gridworld Domain

The third domain that we used was the gridworld domain developed by Thomas (2015b, Section 2.5) for evaluating OPE algorithms. It is a $4 \times 4$ gridworld with four actions, $L = 100$, and deterministic transition and reward functions. This domain was developed specifically for evaluating different OPE methods. Thomas (2015b) proposed five policies, $\pi_1, \ldots, \pi_5$, that can serve as the behavior and evaluation policies.

Although this setup was developed for evaluating OPE methods, it was not developed with DR and WDR in mind (since they were introduced later). Specifically, its use of deterministic state-transition and reward functions means that when the model is accurate, AM, DR, and WDR will all perform similarly (due the the **(b)** term in (3) being near-zero).

We therefore performed experiments with two variants of this gridworld. In the first variant the approximate model was provided with the horizon, $L = 100$. However, in the second variant we introduced some partial observability by providing the model with the incorrect horizon: $L = 101$. This has a significant impact for value predictions close to the end of a trajectory because the model incorrectly predicts when the rewards will necessarily be zero. We write *Gridworld-TH* and *Gridworld-FH* to denote the gridworld where the agent is provided with the true horizon and false horizon, respectively.

### D.4. Experimental Setup

For each domain we generated $n$ trajectories (for various $n$) and computed the sample mean squared error between the predictions of the various OPE methods and the true performance of the evaluation policy (estimated using a large number of on-policy Monte-Carlo rollouts). For each value of $n$ and each OPE algorithm, we performed this experiment 128 times and report the average sample mean squared error over these 128 trials. All plots include standard error bars and use logarithmic scales for both the horizontal and vertical axes.

Perhaps surprisingly, it is not obvious how to fairly compare the different OPE algorithms. Clearly IS, PDIS, WIS, and CWPDIS should use all of the trajectories in $D$, since they do not require an approximate model. Similarly, AM should use all of the data to construct an approximate model. However, how should the available data be split for DR, WDR, and the MAGIC estimators? We believe that there are at least three reasonable answers:

1. DR, WDR, and MAGIC should be provided with additional trajectories not available to IS, PDIS, WIS, and CWPDIS, and these trajectories should be used to construct an approximate model. This setup would emulate the setting where prior domain knowledge (not necessarily trajectories) can be used to construct an approximate model, which IS, PDIS, WIS, and CWPDIS ignore.
2. DR, WDR, and MAGIC should use all of the available data, $D$, to construct an approximate model. They should then reuse this same data to compute their estimates. This approach is reasonable, but the reuse of data invalidates our theoretical guarantees. Still, empirically we find that this approach causes DR, WDR, and MAGIC to perform at their best.
3. DR, WDR, and MAGIC should partition $D$ into two sets. The first set should be used to construct the approximate model, and the second set should be used to compute the DR, WDR, and MAGIC estimates using the approximate model.

Since there is not necessarily a "correct" answer to which way of performing experiments is best, we show our results using both the second and third approach. For each domain, the "full-data" variant uses the second approach while the "half-data" variant uses the third approach, where $D$ is partitioned into two sets of equal size.

Since all of the domains that we use have finite state and action sets, we use a simple maximum-likelihood approximate model. That is, we predict that the probability of transitioning from $s$ to $s'$ given action $a$ is the number of times this transition was observed divided by the number of times action $a$ was taken in state $s$. If $D$ contains no examples of action $a$ being taken in state $s$, then we assume that taking action $a$ in state $s$ always causes a transition to the terminal absorbing state.

In this appendix, we present empirical results from four previous importance sampling methods, definitions of which can be found in the work of Thomas (2015b, Chapter 3): *importance sampling* (IS), *per-decision importance sampling* (PDIS), *weighted importance sampling* (WIS), and *consistent weighted per-decision importance sampling* (CWPDIS). We also show results for the guided importance sampling methods DR and WDR and the purely model-based method, AM. The legend used by all of the plots in this appendix is provided in Figure 5.

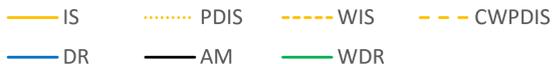

Figure 5: The legend used by all plots in Appendix D.

### D.5. ModelFail Results

Figure 1b in Section 6 depicts the result on the ModelFail domain in the full-data setting. We reproduce this plot in Figure 6. Here the weighted importance sampling methods, WIS and CWPDIS, are obscured by the curve for WDR, while the unweighted importance sampling methods, IS and PDIS, are obscured by the curve for DR. Notice that WDR outperforms AM by orders of magnitude and DR by approximately an order of magnitude. Also notice that even though the approximate model is not accurate, which means that the control variates used by DR and WDR may be poor, the DR and WDR estimators do not perform worse than PDIS and CWPDIS, respectively.

In Figure 7 we reproduce this experiment in the half-data setting. Since AM does not use any data for importance sampling, in both settings (half-data and full-data) it is identical. Similarly, IS, PDIS, WIS, and CWPDIS do not use an approximate model, so they always use all of the data and are therefore also identical in both settings. However, DR and WDR are not the same—they use half of the data to construct the approximate model and the other half to compute their estimates. This means that, for DR and WDR, the approximate model tends to be worse, and the importance sampling estimate also tends to be worse. As a result, the DR and WDR curves are shifted up slightly. Still, the same general trends are evident—WDR outperforms AM by orders and DR by an order.

### D.6. ModelWin Results

Figure 1c in Section 6 depicts the result of running importance sampling and guided importance sampling methods as well as the approximate model estimator on the ModelWin experimental setup in the full-data setting. We reproduce this plot in Figure 8. Here AM has approximately an order of magnitude lower MSE than all of the other methods, including WDR, and was our motivation for combining AM and WDR using BIM.

In Figure 9 we reproduce this experiment in the half-data setting. As with the ModelWin setup, this only hurts DR and WDR. When there are few trajectories, it appears to impact DR more than WDR, although this may be due to noise (notice the large standard error bars on the DR curve when $n$ is small.

### D.7. Gridworld Results

Figure 1a in Section 6 depicts the results of using the fourth gridworld policy, $\pi_4$, as the behavior policy and the fifth, $\pi_5$, as the evaluation policy for the Gridworld-FH domain in the full-data setting. We reproduce it in Figure 10. Notice that WDR outperforms all other methods by at least an order of magnitude.

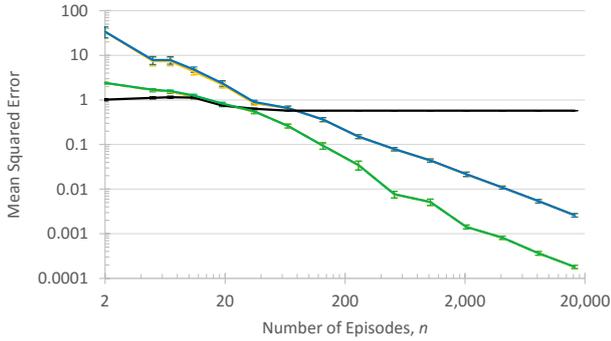

Figure 6: ModelFail, full-data.

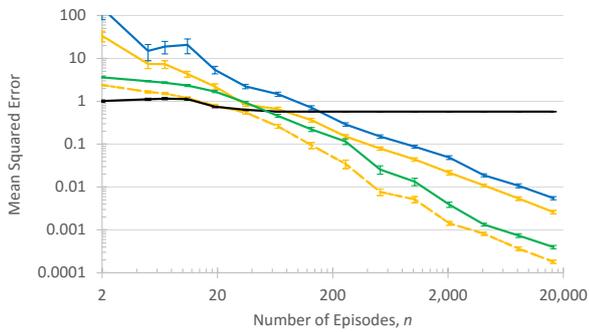

Figure 7: ModelFail, half-data.

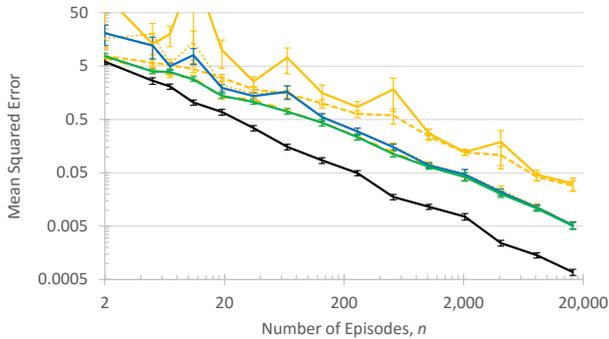

Figure 8: ModelWin, full-data.

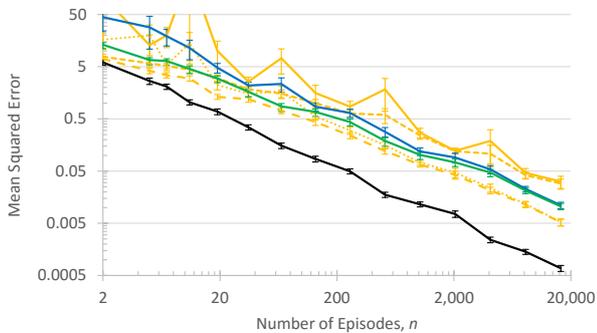

Figure 9: ModelWin, half-data.

In Figure 11 we reproduce this experiment in the half-data setting. As before there is little change, except that the DR and WDR curves shift up. WDR remains the best-performing estimator, by approximately an order of magnitude.

Next we reproduced Figures 10 and 11 for Gridworld-TH as opposed to Gridworld-FH. The results are in Figures 12 and 13 respectively. Notice that, when given the true horizon, AM excels. In the full-data setting DR and WDR both lie directly on top of the curve for AM. This makes sense because the transition function and reward function are deterministic, and so, given the way that we constructed our approximate model, both methods degenerate to exactly AM. In the half-data setting DR and WDR lag slightly behind the curve for AM since they can only use half as much data.

Next we reproduced these four figures using the first gridworld policy, $\pi_1$, as the behavior policy and the second, $\pi_2$, as the evaluation policy. Whereas $\pi_4$ and $\pi_5$ are nearly deterministic and produce long trajectories, $\pi_1$ and $\pi_2$ are far from deterministic and tend to produce shorter trajectories. Notably, the behavior policy, $\pi_1$, selects actions uniformly randomly, and so this presents a very different setting for OPE. The results are provided in Figures 14–17. In this example, DR and WDR perform similarly—significantly better than the importance sampling algorithms IS, PDIS, WIS, and CWPDIS, and marginally better than AM given enough data. Also, when the true horizon is provided to the model, DR and WDR again degenerate to AM.

### D.8. Summary

The key takeaways from these experiments are that WDR tends to outperform the other importance sampling estimators, IS, PDIS, WIS, and CWPDIS, as well as the guided importance sampling method, DR. None of these methods achieved mean squared errors within an order of magnitude of WDR's across all of our experiments. This shows the power of WDR as a guided importance sampling method.

However, WDR did not always win—in the ModelFail setting, AM outperformed WDR by an order of magnitude. Similar results have been observed by others. For example, in the experiments of Jiang & Li (2015), AM tended to outperform DR (although they did not compare to WDR, since it had not yet been introduced). This motivated our introduction of the BIM estimator as a way to blend together WDR and AM.

Notice that, if the transition function and reward function are deterministic and there is no partial observability (as in the gridworld experiments using the true horizon), then, given the way that we constructed our approximate model, DR and WDR degenerate to AM. This degeneration (which

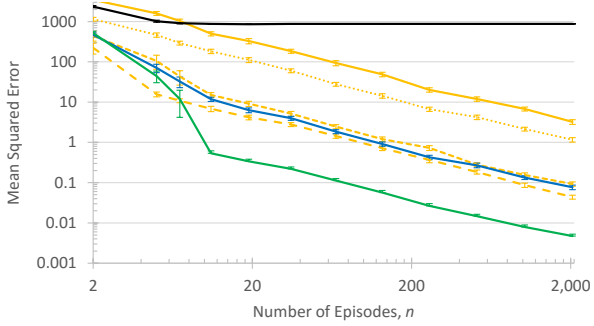

Figure 10: Gridworld-FH, full-data, $\pi_4$ behavior policy, $\pi_5$ evaluation policy.

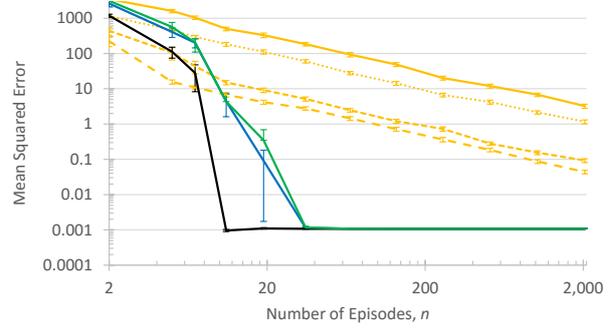

Figure 13: Gridworld-TH, half-data, $\pi_4$ behavior policy, $\pi_5$ evaluation policy.

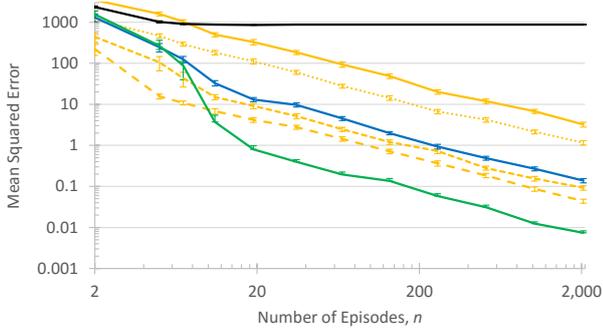

Figure 11: Gridworld-FH, half-data, $\pi_4$ behavior policy, $\pi_5$ evaluation policy.

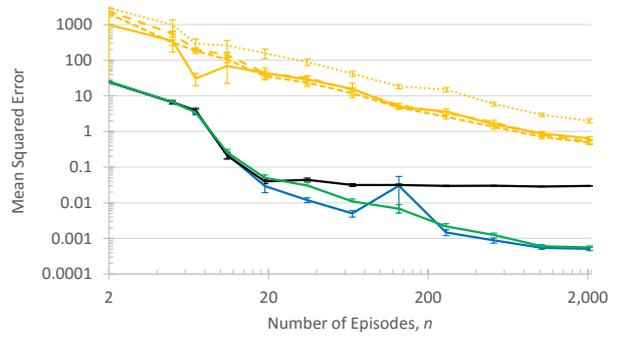

Figure 14: Gridworld-FH, full-data, $\pi_1$ behavior policy, $\pi_2$ evaluation policy.

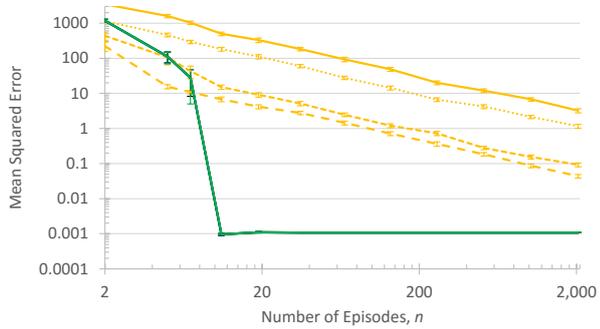

Figure 12: Gridworld-TH, full-data, $\pi_4$ behavior policy, $\pi_5$ evaluation policy.

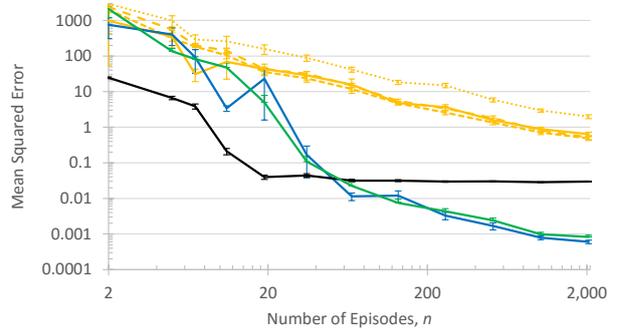

Figure 15: Gridworld-FH, half-data, $\pi_1$ behavior policy, $\pi_2$ evaluation policy.

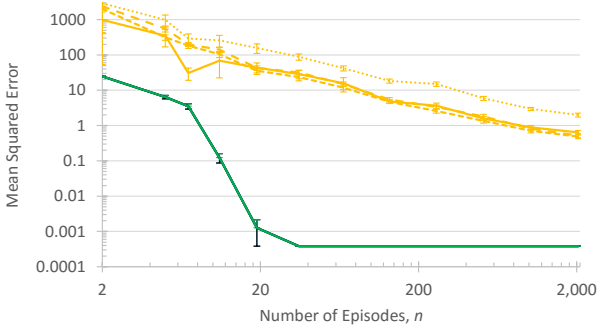

Figure 16: Gridworld-TH, full-data, $\pi_1$ behavior policy, $\pi_2$ evaluation policy.

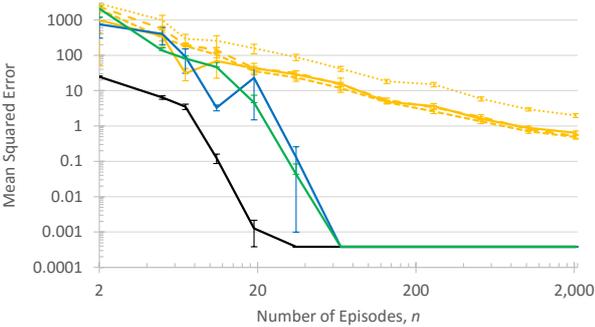

Figure 17: Gridworld-TH, half-data, $\pi_1$ behavior policy, $\pi_2$ evaluation policy.

is not bad, but suggests that importance sampling methods are not necessary) would also not occur if the approximate model used function approximation.

Lastly, notice that DR and WDR performed better in the full-data setting than in the half-data setting. This suggests that, in practice, one should use all of the available data both to produce an approximate model and to compute the DR and WDR estimates. Even though this violates the assumptions used by our theoretical guarantees, this does not mean, for example, that MAGIC will not still be a strongly consistent estimator for the application at hand.

## E. Consistency of BIM

In this appendix we prove Theorem 3, which states that if Assumption 4 holds, there exists at least one $j \in \mathcal{J}$ such that $g^{(j)}(D)$ is a strongly consistent estimator of $v(\pi_e)$, and $\widehat{\mathbf{b}}_n - \mathbf{b}_n \xrightarrow{\text{a.s.}} 0$, and $\widehat{\Omega}_n - \Omega_n \xrightarrow{\text{a.s.}} 0$, then $\text{BIM}(D, \widehat{\Omega}_n, \widehat{\mathbf{b}}_n) \xrightarrow{\text{a.s.}} v(\pi_e)$.

We begin by showing that BIM converges almost surely to $v(\pi_e)$ if it were to use the true $\Omega_n$ and $\mathbf{b}_n$, rather than estimates thereof. Let $j^\star \in \mathcal{J}$ be an index such that $g^{(j^\star)}(D) \xrightarrow{\text{a.s.}} v(\pi_e)$, which exists by assumption. Let $\mathbf{y} \in \Delta^{|\mathcal{J}|}$ be the weight vector that places a weight of one on $g^{(j^\star)}(D)$ and a weight of zero on the other returns, such that $\mathbf{y}^\intercal \mathbf{g}_\mathcal{J}(D) = g^{(j^\star)}(D) \xrightarrow{\text{a.s.}} v(\pi_e)$. So, by Lemma 3 (which requires that $g^{(j)}(D)$ is uniformly bounded for all $j \in \mathcal{J}$, which holds by Assumption 4 and the fact that rewards and reward predictions are bounded), we have that $\lim_{n \to \infty} \text{MSE}(\mathbf{y}^\intercal \mathbf{g}(D), v(\pi_e)) = 0$.

Recall that $\text{BIM}(D, \Omega_n, \mathbf{b}_n)$ uses the weight vector, $\mathbf{x}^\star$ that minimizes the MSE:

$$\mathbf{x}^\star \in \arg\min_{\mathbf{x} \in \Delta^{|\mathcal{J}|}} \text{MSE}(\mathbf{x}^\intercal \mathbf{g}_\mathcal{J}(D), \Omega_n, \mathbf{b}_n).$$

Since $\mathbf{y} \in \Delta^{|\mathcal{J}|}$, we have that for all $n$

$$\text{MSE}((\mathbf{x}^\star)^\intercal \mathbf{g}_\mathcal{J}(D), v(\pi_e)) \leq \text{MSE}(\mathbf{y}^\intercal \mathbf{g}_\mathcal{J}(D), v(\pi_e)).$$

Since $\lim_{n \to \infty} \text{MSE}(\mathbf{y}^\intercal \mathbf{g}_\mathcal{J}(D), v(\pi_e)) = 0$ we have that

$$\lim_{n \to \infty} \text{MSE}((\mathbf{x}^\star)^\intercal \mathbf{g}_\mathcal{J}(D), v(\pi_e)) \leq 0,$$

and since MSE is always greater than or equal to zero, we can replace the $\leq$ above with an equality. Since $(\mathbf{x}^\star)^\intercal \mathbf{g}_\mathcal{J}(D) = \text{BIM}(D, \Omega_n, \mathbf{b}_n)$ this can be rewritten as

$$\lim_{n \to \infty} \text{MSE}(\text{BIM}(D, \Omega_n, \mathbf{b}_n), v(\pi_e)) = 0.$$

By Lemma 3 we have that this implies that $\text{BIM}(D, \Omega_n \mathbf{b}_n) \xrightarrow{\text{a.s.}} v(\pi_e)$.

So far we have shown that BIM, when using the true covariance matrix and bias vector, converges almost surely to $v(\pi_e)$. By Lemma 5 we can therefore conclude that if $\widehat{\mathbf{b}}_n - \mathbf{b}_n \xrightarrow{\text{a.s.}} 0$ and $\widehat{\Omega}_n - \Omega_n \xrightarrow{\text{a.s.}} 0$, then $\text{BIM}(D, \widehat{\Omega}_n \widehat{\mathbf{b}}_n) \xrightarrow{\text{a.s.}} v(\pi_e)$.

# F. Derivation of $g^{(j)}(D)$ using WDR

In this appendix we derive a reasonable definition for $g^{(j)}(D)$, the off-policy $j$-step return, when using WDR for the importance sampling estimator. We assume that the reader is familiar with our use of control variates in Appendix B. First, consider what control variate should be added to the $j$-step PDIS or CWPDIS estimator:

$$\sum_{i=1}^{n} \sum_{t=0}^{j} \gamma^t w_t^i R_t^{H_i},$$

where the definition of $w_t^i$ determines whether this is PDIS or CWPDIS. Reproducing our arguments from Appendix B, we find that a reasonable definition for $\text{IS}^{(j)}(D)$ is similar to (15), but with the time index, $t$, summing only to $t = j$ and using $w_t^i$ terms rather than $\rho_t^i$ terms for generality:

$$\text{IS}^{(j)}(D) \coloneqq \sum_{i=1}^{n} \sum_{t=0}^{j} w_t^i \gamma^t R_t^{H_i}$$
$$- \sum_{i=1}^{n} \sum_{t=0}^{j} \gamma^t \sum_{\tau=0}^{t} w_\tau^i \hat{r}^{\pi_e}\left(S_\tau^{H_i}, A_\tau^{H_i}, t-\tau\right)$$
$$+ \sum_{i=1}^{n} \sum_{t=0}^{j} \gamma^t \sum_{\tau=0}^{t} w_{\tau-1}^i \hat{r}^{\pi_e}\left(S_\tau^{H_i}, t-\tau\right).$$

Notice that this definition is *not* equivalent to what one would get if (2) were modified only so that the sum goes from time $t = 0$ to $t = j$, since that definition would include reward predictions beyond $R_j$ in $\hat{v}$ and $\hat{q}$ terms. Instead, this definition is equivalent to the definition of (2) if it were applied to a modified MDP where every episode terminates after $R_j$ is produced.

Next, consider the definition of $\text{AM}^{(j)}(D)$. We might use importance sampling to correct for the distribution of $S_j$, and the model to predict the remaining rewards:[13]

$$\text{AM}^{(j)}(D) = \gamma^j \sum_{i=1}^{n} w_{j-1}^i \hat{v}^{\pi_e}(S_j^{H_i})$$
$$= \gamma^j \sum_{i=1}^{n} w_{j-1}^i \sum_{\tau=0}^{\infty} \gamma^\tau \hat{r}^{\pi_e}(S_j^{H_i}, \tau).$$

---

[13]This is just one possible definition of $\text{AM}^{(j)}$. We also experimented with a definition that is purely model based: $\text{AM}^{(j)}(D) \coloneqq \sum_{s \in \mathcal{S}} \widehat{d}_0(s) \sum_{t=j}^{\infty} \gamma^t \hat{r}^{\pi_e}(s, t)$. Since this definition does not include any importance weights, it does not require an additional control variate. We found that this variant performed similarly to the definition that we present.

Notice that $\text{AM}^{(j)}$ is not a purely model-based estimator if $j \geq 0$ since it uses importance weights. Furthermore, this use of importance sampling can result in high variance. To partially mitigate this variance, we can introduce a control variate to get a new definition:

$$\text{AM}^{(j)}(D) = \gamma^j \sum_{i=1}^{n} w_{j-1}^i \sum_{\tau=0}^{\infty} \gamma^\tau \hat{r}^{\pi_e}(S_j^{H_i}, \tau)$$
$$- \gamma^j \sum_{i=1}^{n} w_{j-1}^i \sum_{\tau=0}^{\infty} \gamma^\tau \hat{r}^{\pi_e}(S_{j-1}^{H_i}, A_{j-1}^{H_i}, \tau+1)$$
$$+ \gamma^j \sum_{i=1}^{n} w_{j-2}^i \sum_{\tau=0}^{\infty} \gamma^\tau \hat{r}^{\pi_e}(S_{j-1}^{H_i}, \tau+1).$$

As in our derivation of the DR estimator in Appendix B, we can repeat this process by continuing to add control variates until the control variate is not random to get our final definition of $\text{AM}^{(j)}(D)$:

$$\text{AM}^{(j)}(D) \coloneqq \gamma^j \sum_{i=1}^{n} w_{j-1}^i \sum_{\tau=0}^{\infty} \gamma^\tau \hat{r}^{\pi_e}(S_j^{H_i}, \tau)$$
$$- \gamma^j \sum_{k=1}^{j} \sum_{i=1}^{n} w_{j-k}^i \sum_{\tau=0}^{\infty} \gamma^\tau \hat{r}^{\pi_e}(S_{j-k}^{H_i}, A_{j-k}^{H_i}, \tau+k)$$
$$+ \gamma^j \sum_{k=1}^{j} \sum_{i=1}^{n} w_{j-k-1}^i \sum_{\tau=0}^{\infty} \gamma^\tau \hat{r}^{\pi_e}(S_{j-k}^{H_i}, \tau+k).$$

Combining the IS and AM definitions to produce a off-policy $j$-step return as defined in (4) we have:

$$g^{(j)}(D) \coloneqq \text{IS}^{(j)}(D) + \text{AM}^{(j+1)}(D)$$
$$= \sum_{i=1}^{n} \sum_{t=0}^{j} w_t^i \gamma^t R_t^{H_i} + \gamma^{j+1} \sum_{i=1}^{n} w_j^i \sum_{\tau=0}^{\infty} \hat{r}^{\pi_e}(S_{j+1}^{H_i}, \tau)$$
$$- \underbrace{\sum_{i=1}^{n} \sum_{t=0}^{j} \gamma^t \sum_{\tau=0}^{t} w_\tau^i \hat{r}^{\pi_e}\left(S_\tau^{H_i}, A_\tau^{H_i}, t-\tau\right)}_{(a)}$$
$$+ \underbrace{\sum_{i=1}^{n} \sum_{t=0}^{j} \gamma^t \sum_{\tau=0}^{t} w_{\tau-1}^i \hat{r}^{\pi_e}\left(S_\tau^{H_i}, t-\tau\right)}_{(b)}$$
$$- \underbrace{\gamma^{j+1} \sum_{k=1}^{j+1} \sum_{i=1}^{n} w_{j+1-k}^i \sum_{\tau=0}^{\infty} \gamma^\tau \hat{r}^{\pi_e}(S_{j+1-k}^{H_i}, A_{j+1-k}^{H_i}, \tau+k)}_{(c)}$$
$$+ \underbrace{\gamma^{j+1} \sum_{k=1}^{j+1} \sum_{i=1}^{n} w_{j-k}^i \sum_{\tau=0}^{\infty} \gamma^\tau \hat{r}^{\pi_e}(S_{j+1-k}^{H_i}, \tau+k)}_{(d)}.$$

Notice that the terms **(a)** and **(b)** use predictions of rewards up until and including $R_j$, while the terms **(c)** and **(d)** use predictions of rewards beginning with $R_{j+1}$ and going to infinity. So, with algebraic manipulations we can combine **(a)** and **(c)** to get

$$\sum_{i=1}^{n}\sum_{t=0}^{j}\gamma^t w_t^i \hat{q}^{\pi_e}\left(S_t^{H_i}, A_t^{H_i}\right)$$

and we can combine **(b)** and **(d)** to get:

$$\sum_{i=1}^{n}\sum_{t=0}^{j}\gamma^t w_{t-1}^i \hat{v}^{\pi_e}\left(S_t^{H_i}\right).$$

So, we have that

$$g^{(j)}(D) := \sum_{i=1}^{n}\sum_{t=0}^{j}\gamma^t w_t^i R_t^{H_i} + \sum_{i=1}^{n}\gamma^{j+1} w_j^i \hat{v}^{\pi_e}(S_{j+1}^{H_i})$$
$$- \sum_{i=1}^{n}\sum_{t=0}^{j}\gamma^t \left(w_t^i \hat{q}^{\pi_e}\left(S_t^{H_i}, A_t^{H_i}\right) - w_{t-1}^i \hat{v}^{\pi_e}\left(S_t^{H_i}\right)\right).$$

## G. MAGIC Details

In this section we provide additional details about the MAGIC algorithm. Specifically, we describe exactly how we estimate $\Omega_n$ and $\mathbf{b}_n$ before presenting pseudocode for MAGIC.

### G.1. Estimating $\Omega_n$

We can write $g^{(j)}(D)$ as the sum of $n$ terms:

$$g^{(j)}(D) = \sum_{i=1}^{n} g_i^{(j)}(D), \quad (24)$$

where

$$g_i^{(j)}(D) := \left(\sum_{t=0}^{j}\gamma^t w_t^i R_t^{H_i}\right) + \gamma^{j+1} w_j^i \hat{v}^{\pi_e}(S_{j+1}^{H_i})$$
$$- \sum_{t=0}^{j}\gamma^t \left(w_t^i \hat{q}^{\pi_e}\left(S_t^{H_i}, A_t^{H_i}\right) - w_{t-1}^i \hat{v}^{\pi_e}\left(S_t^{H_i}\right)\right).$$

So,

$$\text{Cov}(g^{(i)}(D), g^{(j)}(D)) = \text{Cov}\left(\sum_{k=1}^{n} g_k^{(i)}(D), \sum_{k=1}^{n} g_k^{(j)}(D)\right).$$

Notice that $g_i^{(j)}(D)$ really is a function of all of $D$, not just $H_i$, since $w_t^i = \rho_t^i / \sum_{j=1}^{n}\rho_t^j$. This means that, although the terms in the sum, $\sum_{k=1}^{n} g_k^{(i)}(D)$, are identically distributed, they are not independent, due to their shared reliance on $D$. However, notice that the $g_k^{(i)}(D)$ terms, for various $k$, become less dependent as $n \to \infty$ because the only dependence of $g_k^{(i)}(D)$ on trajectories other than $H_k$ comes from the denominator of $w_t^i$, which converges almost surely to $n$ (we established this in our proofs that WDR is strongly consistent).

We therefore propose an approximation of $\Omega_n$ that comes from the assumption that the $g_k^{(i)}(D)$ terms, for various $k$, are independent:

$$\text{Cov}(g^{(i)}(D), g^{(j)}(D))$$
$$= \sum_{k\in\{1,\ldots,n\}}\sum_{l\in\{1,\ldots,n\}}\text{Cov}\left(g_k^{(i)}(D), g_l^{(j)}(D)\right)$$
$$\overset{(a)}{\approx} \sum_{k\in\{1,\ldots,n\}}\text{Cov}\left(g_k^{(i)}(D), g_k^{(j)}(D)\right)$$
$$\overset{(b)}{=} n\,\text{Cov}\left(g_{(\cdot)}^{(i)}(D), g_{(\cdot)}^{(j)}(D)\right),$$

where **(a)** comes from the assumption that $g_k^{(i)}(D)$ and $g_l^{(j)}(D)$ are independent for all $i, j, k$, and $l$ where $k \neq l$, **(b)** comes from the assumption that they are identically distributed, and where $g_{(\cdot)}^{(i)}(D)$ uses $(\cdot)$ to denote that any subscript in $\{1, \ldots, n\}$ could be used since the random variables are independent and identically distributed.

We therefore approximate $\Omega_n$ using the sample covariance:

$$\widehat{\Omega}_n(i,j) := \frac{n}{n-1}\sum_{k=1}^{n}\left(g_k^{(\mathcal{J}_i)}(D) - \bar{g}_k^{(\mathcal{J}_i)}(D)\right) \quad (25)$$
$$\times \left(g_k^{(\mathcal{J}_j)}(D) - \bar{g}_k^{(\mathcal{J}_j)}(D)\right),$$

where

$$\bar{g}_k^{(\mathcal{J}_i)}(D) := \frac{1}{n}\sum_{k=1}^{n} g_k^{(\mathcal{J}_i)}(D).$$

The above scheme for estimating $\Omega_n$ is the one that we use in our pseudocode and experiments. However, we also experimented with bootstrap estimates of $\Omega_n$. They yielded similar performance at significantly higher computational cost.

### G.2. Estimating $\mathbf{b}_n$

As described previously, we use a confidence interval, $\text{CI}(g^{(\infty)}(D), \delta)$, when computing $\mathbf{b}_n$. We stated that the confidence interval that we use is a combination of the percentile bootstrap and the Chernoff-Hoeffding inequality. Specifically, we compute the confidence interval produced by both methods, and return the *tighter* of the two. In practice, this is nearly always the confidence interval produced by the percentile bootstrap, and so practical implementations of MAGIC may just use the percentile bootstrap. We include the loose Chernoff-Hoeffding bound be-cause it allows for easier theoretical analysis of the MAGIC algorithm.

### G.3. Pseudocode

Pseudocode for the MAGIC algorithm is provided in Algorithm 1. It takes as input $D$, $\pi_e$, and an approximate model, all of which are defined in Section 2. It also takes as input $\mathcal{J}$, which is defined in Section 7, and a positive integer $\kappa$, that we have not defined previously. We use $\kappa$ to denote the number of times the bootstrap algorithm should resample the trajectories. In our experiments we used $\kappa = 200$. In general, it should be made as large as possible given any runtime constraints. Other literature has suggested that it should be chosen to be approximately $\kappa = 2000$ (Efron & Tibshirani, 1993; Davison & Hinkley, 1997).

Line 2 calls for the $|\mathcal{J}| \times |\mathcal{J}|$ matrix, $\widehat{\Omega}_n$, to be computed according to (25).

Line 3 specifies that a structure, $D$, should be created. This structure will be used to store the bootstrap resamplings, such that $D_i$ is the $i^{\text{th}}$ resampling of $D$. That is, $D_i$ is a set of $n$ trajectories and the behavior policies that generated them, sampled with replacement from $D$ (this resampling is done on lines 4–6).

Line 7 calls for the creation of a vector, $\mathbf{v}$, to store the off-policy $j$-step return for $j = \infty$ (recall that this is just the WDR estimator) for each bootstrap sample, sorted into ascending order. Lines 8 and 9 then compute the percentile bootstrap $10\%$ confidence interval, $[l, u]$, for the mean of $g^{(\infty)}(D)$, which we ensure includes $\text{WDR}(D)$. For our theoretical analysis, we add a line after this that sets

$$l \leftarrow \max\left\{l, \text{WDR}(D) - \xi\sqrt{\frac{\ln(2/\delta)}{2n}}\right\} \quad (26)$$

and

$$u \leftarrow \min\left\{l, \text{WDR}(D) + \xi\sqrt{\frac{\ln(2/\delta)}{2n}}\right\}, \quad (27)$$

where $\xi$ is a bound on the range of $g^{(i)}(D)$. In practice, these lines almost never change the values of $l$ and $u$ and can be ignored.

Lines 10–12 then show how the bias vector can be computed from the already defined terms. Notice that the order of $g^{(\mathcal{J}_j)}(D)$ and $l$ or $u$ does not matter since the bias term in the decomposition of mean squared error is squared. The order that we use facilitates a simple consistency proof for MAGIC. Given that the covariance matrix and bias vector have been approximated, Line 13 sets $\mathbf{x}$ to be the solution of a constrained quadratic program (in our experiments we solved this quadratic program using the Gurobi library). Finally, line 14 returns the weighted combination of the different off-policy $j$-step returns (recall that $\mathbf{g}_{\mathcal{J}}(D)$ is defined in Section 7).

---

**Algorithm 1** MAGIC($D$)

1: **Input:**
   - $\mathcal{D}$: Historical data.
   - $\pi_e$: Evaluation policy.
   - Approximate model that allows for computation of $\hat{r}^{\pi_e}(s, a, t)$.
   - $\mathcal{J}$: The set of return lengths to consider. The first element, $\mathcal{J}_1$, should be $-1$ and the last, $\mathcal{J}_{|\mathcal{J}|}$, should be $\infty$.
   - $\kappa$: The number of bootstrap resamplings.
2: Compute $\widehat{\Omega}_n$ according to (25).
3: Allocate $D_{(\cdot)}$ so that for all $i \in \{1, \ldots, \kappa\}$, $D_i$ can hold $n$ trajectories.
4: **for** $i = 1$ **to** $\kappa$ **do**
5:   Load $D_i$ with $n$ uniform random samples drawn from $D$ with replacement.
6: **end for**
7: $\mathbf{v} = \text{sort}\left(g^{(\infty)}(D_{(\cdot)})\right)$
8: $l \leftarrow \min\{\text{WDR}(D), \mathbf{v}(\lfloor 0.05n \rfloor)\}$
9: $u \leftarrow \max\{\text{WDR}(D), \mathbf{v}(\lceil 0.95n \rceil)\}$
10: **for** $j = 1$ **to** $|\mathcal{J}|$ **do**
11:
$$\widehat{\mathbf{b}}_n(j) \leftarrow \begin{cases} g^{(\mathcal{J}_j)}(D) - u & \text{if } g^{(\mathcal{J}_j)}(D) > u \\ g^{(\mathcal{J}_j)}(D) - l & \text{if } g^{(\mathcal{J}_j)}(D) < l \\ 0 & \text{otherwise.} \end{cases}$$
12: **end for**
13: $\mathbf{x} \leftarrow \arg\min_{\mathbf{x} \in \Delta^{|\mathcal{J}|}} \mathbf{x}^\intercal [\widehat{\Omega}_n + \widehat{\mathbf{b}}_n \widehat{\mathbf{b}}_n^\intercal] \mathbf{x}$
14: **return** $\mathbf{x}^\intercal \mathbf{g}_{\mathcal{J}}(D)$

---

## H. Consistency of MAGIC

In this section we prove Theorem 4, which states that if Assumptions 1 and 4 hold and $\infty \in \mathcal{J}$, then $\text{MAGIC}(D) \xrightarrow{\text{a.s.}} v(\pi_e)$. This result follows immediately from Theorem 3 if $\widehat{\Omega}_n - \Omega_n \xrightarrow{\text{a.s.}} 0$ and $\widehat{\mathbf{b}}_n - \mathbf{b}_n \xrightarrow{\text{a.s.}} 0$, since Assumptions 1 and 4 are sufficient to ensure that $g^{(\infty)}(D) = \text{WDR}(D) \xrightarrow{\text{a.s.}} v(\pi_e)$. In Appendix H.3 we show that $\widehat{\Omega}_n - \Omega_n \xrightarrow{\text{a.s.}} 0$, and then in Appendix H.4 we show that $\widehat{\mathbf{b}}_n - \mathbf{b}_n \xrightarrow{\text{a.s.}} 0$. However, first we establish two useful properties of the off-policy $j$-step returns.

### H.1. Convergence of Off-Policy $j$-Step Return

Recall that the off-policy $j$-step return used by MAGIC is given by:

$$g^{(j)}(D) := \sum_{i=1}^{n}\sum_{t=0}^{j} \gamma^t w_t^i R_t^{H_i} + \sum_{i=1}^{n} \gamma^{j+1} w_j^i \hat{v}^{\pi_e}(S_{j+1}^{H_i})$$
$$- \sum_{i=1}^{n}\sum_{t=0}^{j} \gamma^t \left( w_t^i \hat{q}^{\pi_e}\left(S_t^{H_i}, A_t^{H_i}\right) - w_{t-1}^i \hat{v}^{\pi_e}\left(S_t^{H_i}\right) \right),$$

which can be written as:

$$g^{(j)}(D) = \sum_{i=1}^{n}\sum_{t=0}^{j} \gamma^t w_t^i X_t^i + \frac{1}{n}\sum_{i=1}^{n} \hat{v}^{\pi_e}(S_0^{H_i}),$$

where

$$X_t^i = R_t^{H_i} - \hat{q}^{\pi_e}\left(S_t^{H_i}, A_t^{H_i}\right) + \gamma \hat{v}^{\pi_e}\left(S_{t+1}^{H_i}\right). \quad (28)$$

Notice that $X_t^i$ is a bounded random variable since rewards and reward predictions are bounded. So, by Lemma 12 we have that

$$\sum_{i=1}^{n}\sum_{t=0}^{j} \gamma^t w_t^i X_t^i \xrightarrow{\text{a.s.}} \mathbf{E}\left[\sum_{t=0}^{j} \gamma^t X_t \,\middle|\, H \sim \pi_e\right]. \quad (29)$$

Also, since $\hat{v}^{\pi_e}(S_0^{H_i})$ is bounded, we have from the Kolmogorov strong law of large numbers that

$$\frac{1}{n}\sum_{i=1}^{n} \hat{v}^{\pi_e}(S_0^{H_i}) \xrightarrow{\text{a.s.}} \mathbf{E}[\hat{v}^{\pi_e}(S_0)]. \quad (30)$$

So, combining (29) and (30) we have from Property 3 that

$$g^{(j)}(D) \xrightarrow{\text{a.s.}} \mathbf{E}\left[\hat{v}^{\pi_e}(S_0^H) + \sum_{t=0}^{j} \gamma^t X_t \,\middle|\, H \sim \pi_e\right].$$

Let $c_j := \mathbf{E}\left[\hat{v}^{\pi_e}(S_0^H) + \sum_{t=0}^{j} \gamma^t X_t \,\middle|\, H \sim \pi_e\right]$ denote this constant value that $g^{(j)}(D)$ converges to.

### H.2. Convergence of Component of Off-Policy $j$-Step Return

Recall from (24) that the off-policy $j$-step return can be written as:

$$g^{(j)}(D) = \sum_{i=1}^{n} g_i^{(j)}(D),$$

where

$$g_i^{(j)}(D) := \left(\sum_{t=0}^{j} \gamma^t w_t^i R_t^{H_i}\right) + \gamma^{j+1} w_j^i \hat{v}^{\pi_e}(S_{j+1}^{H_i})$$
$$- \sum_{t=0}^{j} \gamma^t \left(w_t^i \hat{q}^{\pi_e}\left(S_t^{H_i}, A_t^{H_i}\right) - w_{t-1}^i \hat{v}^{\pi_e}\left(S_t^{H_i}\right)\right).$$

here we will show that for any $i$ and $j$, $g_i^{(j)}(D) \xrightarrow{\text{a.s.}} 0$.

Notice that $g_i^{(j)}(D)$ can be written as:

$$g_i^{(j)}(D) = \sum_{t=0}^{j} \gamma^t \frac{\rho_t^i X_t^i}{\sum_{k=1}^{n} \rho_t^k}$$
$$= \sum_{t=0}^{j} \gamma^t Y_t^i,$$

where $X_t^i$ is as defined in (28), and

$$Y_t^i := \frac{\rho_t^i X_t^i}{\sum_{k=1}^{n} \rho_t^k}$$
$$= \frac{\frac{1}{n}\rho_t^i X_t^i}{\frac{1}{n}\sum_{k=1}^{n} \rho_t^k}$$

Since $X_t^i$ and $\rho_t^i$ are bounded, we have that $\lim_{n\to\infty} \frac{1}{n}\rho_t^i X_t^i = 0$. Also, by Lemma 11 and Kolmogorov's strong law of large numbers, we have that $\frac{1}{n}\sum_{k=1}^{n} \rho_t^k \xrightarrow{\text{a.s.}} 1$. So, $Y_t^i \xrightarrow{\text{a.s.}} 0$ for all $t$ and $i$. Furthermore, $Y_t^i$ is bounded since $0 \le \frac{\rho_t^i}{\sum_{k=1}^{n} \rho_t^k} \le 1$ and $X_t^i$ is bounded. So, by Property 4, we have that $g_i^{(j)}(D) \xrightarrow{\text{a.s.}} 0$.

### H.3. Consistency of $\widehat{\Omega}_n$

Here we establish that $\widehat{\Omega}_n - \Omega_n \xrightarrow{\text{a.s.}} 0$. There are two steps to this result. First we will show that $\lim_{n\to\infty} \Omega_n = 0$—the true covariance matrix converges to the zero matrix. We then show that $\widehat{\Omega}_n \xrightarrow{\text{a.s.}} 0$ as well, which means that $\widehat{\Omega}_n - \Omega_n \xrightarrow{\text{a.s.}} 0$.

Recall from Appendix H.1 that $g^{(j)}(D) \xrightarrow{\text{a.s.}} c_j$. We can write

$$\Omega_n(i,j) = \mathbf{E}\left[(g^{(i)}(D) - \mathbf{E}[g^{(i)}(D)])(g^{(j)}(D) - \mathbf{E}[g^{(j)}(D)])\right]$$
$$= \mathbf{E}[Y_n], \quad (31)$$

where

$$Y_n := \left(g^{(i)}(D) - \mathbf{E}[g^{(i)}(D)]\right)\left(g^{(j)}(D) - \mathbf{E}[g^{(j)}(D)]\right).$$

Recall that $g^{(j)}(D) \xrightarrow{\text{a.s.}} c_j$. By Lemma 2 we therefore have that for all $j$, $\lim_{n\to\infty} \mathbf{E}[g^{(j)}(D)] = c_j$. So, by the continuous mapping theorem,

$$Y_n \xrightarrow{\text{a.s.}} (c_i - c_i)(c_j - c_j)$$
$$= 0.$$

So, by applying Lemma 2 to (31) we have that $\lim_{n\to\infty} \Omega_n(i,j) = \lim_{n\to\infty} \mathbf{E}[Y_n] = 0$.

Next we show that $\widehat{\Omega}_n \xrightarrow{\text{a.s.}} 0$. First, recall from Appendix H.2 that for all $j \in \mathcal{J}$ and $k \in \{1, \ldots, n\}$,

$$g_k^{(j)}(D) \xrightarrow{\text{a.s.}} 0.$$

So, by Property 3 we have that $\bar{g}_k^{(j)}(D) \xrightarrow{\text{a.s.}} 0$ as well. So, $g_k^{(j)}(D) - \bar{g}_k^{(j)}(D) \xrightarrow{\text{a.s.}} 0$, and so by Property 3 and the definition of $\widehat{\Omega}_n$, we have that

$$\widehat{\Omega}_n(i,j) \xrightarrow{\text{a.s.}} 0$$

for all $(i,j) \in \mathcal{J}^2$.

### H.4. Consistency of $\widehat{\mathbf{b}}_n$

Here we show that $\widehat{\mathbf{b}}_n - \mathbf{b}_n \xrightarrow{\text{a.s.}} 0$. We have from the definitions of $\widehat{\mathbf{b}}_n$, $l$, and $u$ that:

$$\widehat{\mathbf{b}}_n(j) - \mathbf{b}_n(j) \leq g^{(\mathcal{J}_j)}(D) - l - \mathbf{E}[g^{(\mathcal{J}_j)}(D)] + v(\pi_e). \tag{32}$$

and

$$\widehat{\mathbf{b}}_n(j) - \mathbf{b}_n(j) \geq g^{(\mathcal{J}_j)}(D) - u - \mathbf{E}[g^{(\mathcal{J}_j)}(D)] + v(\pi_e). \tag{33}$$

We will show that both of the right hand sides above converge almost surely to zero, which, by Lemma 4, implies that $\widehat{\mathbf{b}}_n(j) - \mathbf{b}_n(j)$ converges almost surely to zero as well.

First consider (32). We have from Appendix H.1 that **1)** $g^{(\mathcal{J}_j)}(D) \xrightarrow{\text{a.s.}} c_{\mathcal{J}_j}$. So, by Lemma 2 we have that **2)** $\lim_{n \to \infty} \mathbf{E}[g^{(\mathcal{J}_j)}(D)] = \mathbf{E}[c_{\mathcal{J}_j}] = c_{\mathcal{J}_j}$. We also have that $u - l \leq \frac{1}{\sqrt{n}}\sqrt{2\xi^2 \ln(2/\delta)}$, by (26) and (27). Since $\text{WDR}(D) \in [l,u]$, we have that

$$|\text{WDR}(D) - l| \leq \frac{1}{\sqrt{n}}\sqrt{2\xi^2 \ln(2/\delta)}.$$

Since $\xi$ is a constant, the right side is a sequence of constants (not random variables) that converges to zero. The left side is positive and less than the right, and so it too must converge (surely, not just almost surely) to zero: $\lim_{n \to \infty} |\text{WDR}(D) - l| = 0$. So,

$$\Pr(\lim_{n \to \infty} l = v(\pi_e)) = \Pr(\lim_{n \to \infty} l + \text{WDR}(D) - l = v(\pi_e))$$

$$= \Pr(\lim_{n \to \infty} \text{WDR}(D) = v(\pi_e))$$

$$= 1,$$

where the last step comes from Theorem 2. This means that **3)** $l \xrightarrow{\text{a.s.}} v(\pi_e)$.

Combining **1)**, **2)**, and **3)**, we have that the right side of (32) converges almost surely to zero. This same argument, using the upper bound, $u$, rather than the lower bound, $l$, shows that the right side of (33) converges almost surely to zero as well, and so we can conclude.

## I. Extended Empirical Studies (MAGIC)

Here we present detailed results concerning the MAGIC estimator. These results will use the same three domains and two experimental setups (full-data and half-data) that were introduced in Appendix D, as well as one additional domain, which we call the *Hybrid* domain. We begin by introducing the Hybrid domain, we then discuss minor changes to the experimental setup and then present results.

### I.1. The Hybrid Domain

The purpose of this domain is to showcase a common problem type: domains where early in a trajectory there is partial observability, but as time passes within each trajectory, the partial observability decays. This happens, for example, in robotics applications where there may be some uncertainty about the position or pose of a robot. However, as the trajectory progresses the robot may be able to better localize itself, removing or diminishing the uncertainty.

We emulate this setting by concatenating the ModelFail and ModelWin domains. That is, the agent begins in the ModelFail domain. Whenever it would transition to the absorbing state, it instead transitions to the initial state of the ModelWin domain.

### I.2. Experimental Setup

We performed these experiments in the same way as those in Appendix D, except that we compared different estimators. Specifically, we introduce curves for the MAGIC estimator, but remove the curves for the poorly-performing importance sampling estimators, IS, PDIS, WIS, and CW-PDIS. So, the plots contain curves for DR, WDR, AM, and MAGIC. The legend used by all of the plots in this appendix is provided in Figure 18.

——— DR ——— AM ——— WDR ——— MAGIC ········ MAGIC-B

Figure 18: The legend used by all plots in Appendix I.

Also, for the hybrid domain we included a curve for *binary MAGIC* (MAGIC-B), which uses $\mathcal{J} = \{-1, \infty\}$. Whereas MAGIC blends between AM and WDR using off-policy $j$-step returns of various lengths, binary MAGIC only places weights on AM and WDR. Our comparison to MAGIC-B shows the importance of including the off-policy $j$-step returns rather than merely trying to switch between, or directly weight, AM and WDR.

Lastly, since all of the domains have finite horizons, we used $\mathcal{J} = \{-1, \ldots, L\}$ for MAGIC. This means that it uses all of the possible off-policy $j$-step returns.

### I.3. ModelFail Results

Figure 2b in Section 9 depicts the results for the ModelFail domain in the full-data setting. We reproduce this plot in Figure 19. In Figure 20 we show the results for ModelFail in the half-data setting. There is little difference between the plots—in both cases MAGIC properly tracks WDR, so that both WDR and MAGIC outperform AM an DR by at least an order of magnitude for most $n$.

### I.4. ModelWin Results

Figure 2c in Section 9 depicts the results for the ModelWin domain in the full-data setting. We reproduce this plot in Figure 21. In Figure 22 we show the results for ModelFail in the half-data setting. In both cases MAGIC tracks AM, although it drifts away a little as $n$ increases. This suggests that there may be room for improvement in our estimates of $\Omega_n$ and $\mathbf{b}_n$. However, also notice that due to the logarithmic scale, the difference between MAGIC and AM is small in comparison to the distance between MAGIC and DR.

### I.5. Gridworld Results

Figures 23 through 30 depict the results for the Gridworld-FH and Gridworld-TH domains in both the full and half-data settings. The same general trends are visible. First, WDR tends to outperform DR, sometimes by an order of magnitude. Also, MAGIC tends to track WDR, since in these experiments it is usually the best-performing algorithm. Lastly, for the Gridworld-TH, full-data setting, DR, WDR, and MAGIC all degenerate to AM, while in the Gridworld-TH, half-data setting they degenerate to approximately AM using half as much data.

### I.6. Hybrid Results

Last, but not least, Figures 31 and 32 show the results on the Hybrid domain in the full-data and half-data settings, respectively. Notice that in MAGIC significantly outperforms all other methods, including WDR and AM. MAGIC also outperforms MAGIC-B, which shows the importance of using off-policy $j$-step returns for various values of $j$.

### I.7. Summary

Overall, MAGIC acts as desired—it tracks WDR or AM, whichever is better for the application at hand. However, notice that it does not do this perfectly, particularly when there is little data available. This is likely because when there is little data it is difficult to estimate $\Omega_n$, and the confidence interval used when estimating $\mathbf{b}_n$ will be loose. In some cases, even when there is a large amount of data, MAGIC struggles to properly track AM. However, this tends to be when both methods perform well, and may be

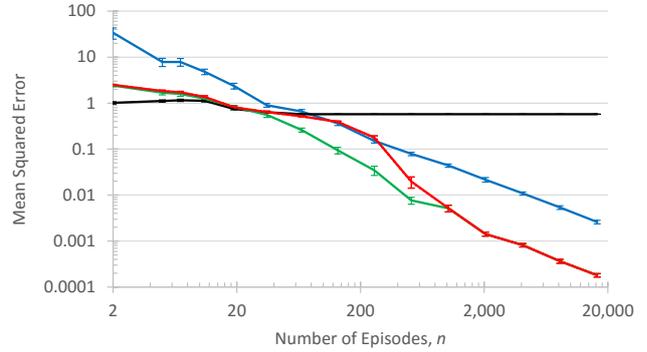

Figure 19: ModelFail, full-data.

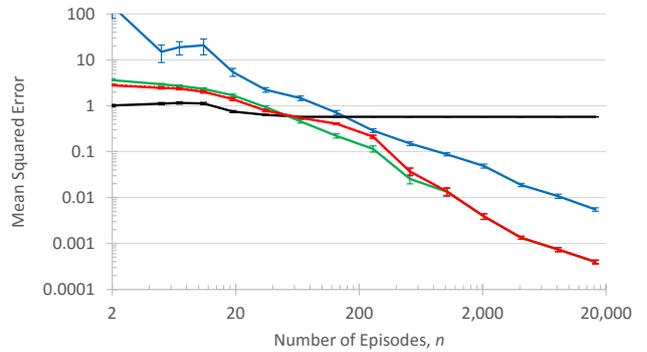

Figure 20: ModelFail, half-data.

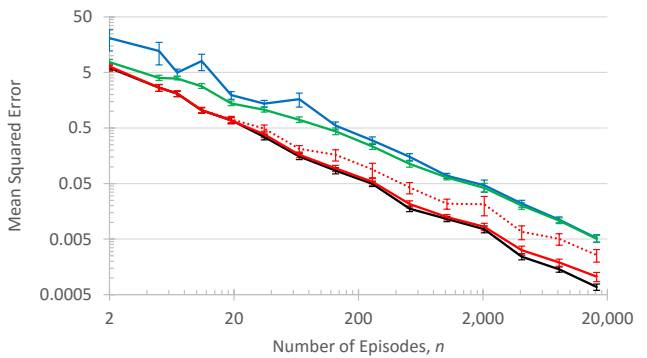

Figure 21: ModelWin, full-data.

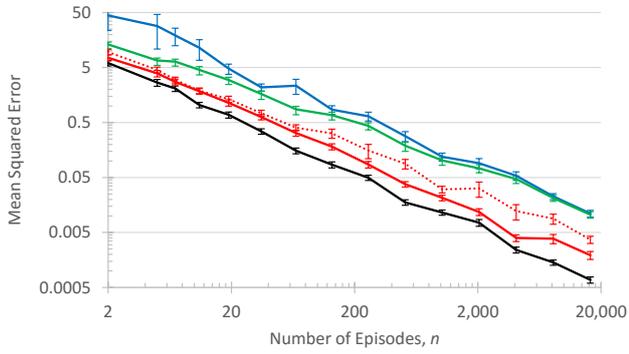

Figure 22: ModelWin, half-data.

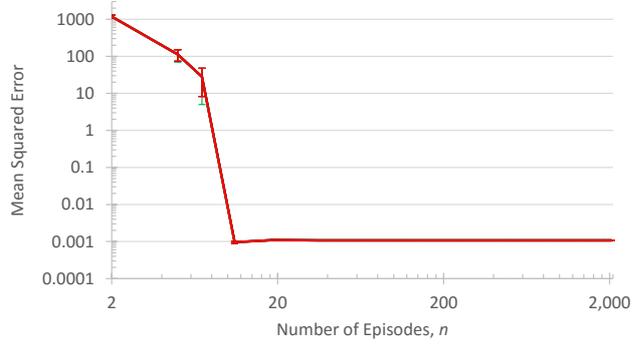

Figure 25: Gridworld-TH, full-data.

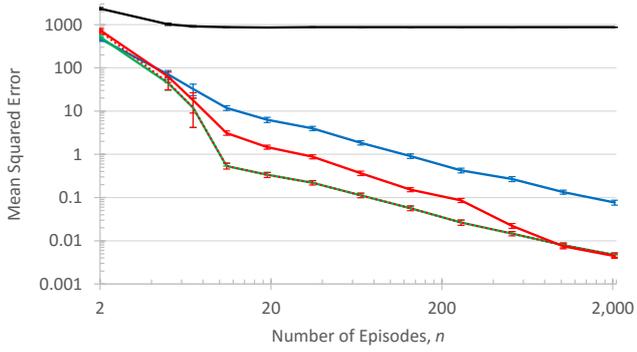

Figure 23: Gridworld-FH, full-data.

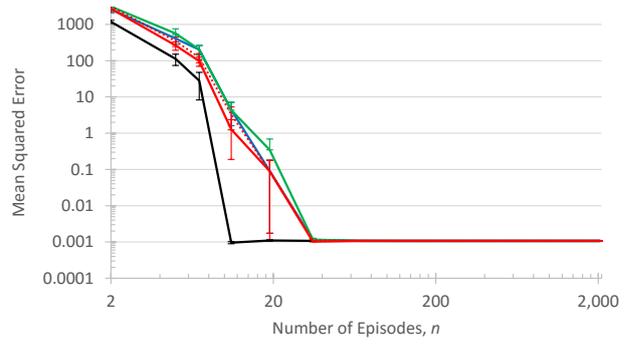

Figure 26: Gridworld-TH, half-data.

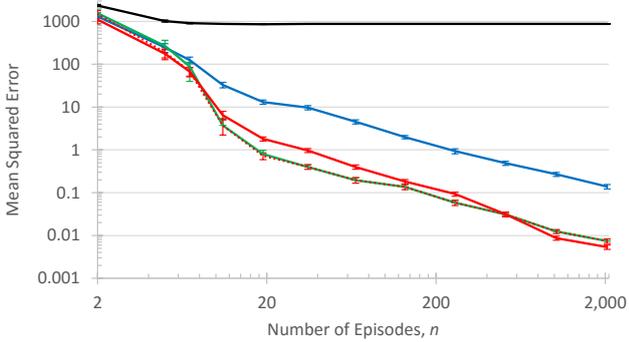

Figure 24: Gridworld-FH, half-data.

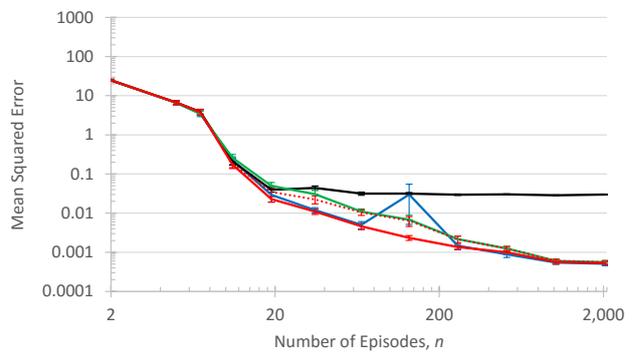

Figure 27: Gridworld-FH, full-data. p1p2

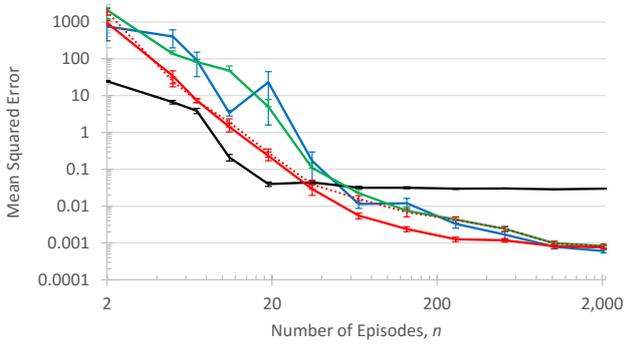

Figure 28: Gridworld-FH, half-data. p1p2

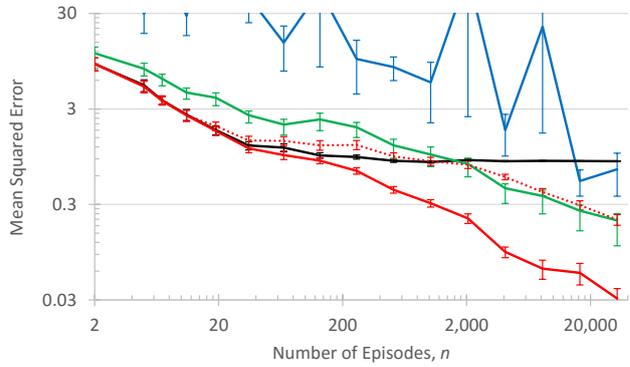

Figure 31: Hybrid, full-data.

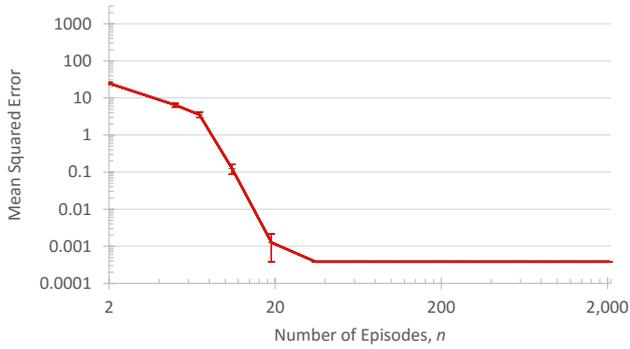

Figure 29: Gridworld-TH, full-data. p1p2

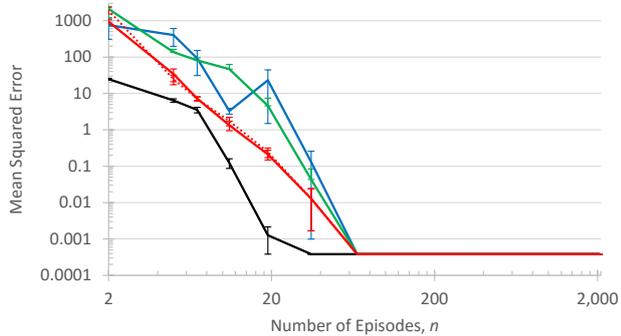

Figure 30: Gridworld-TH, half-data. p1p2

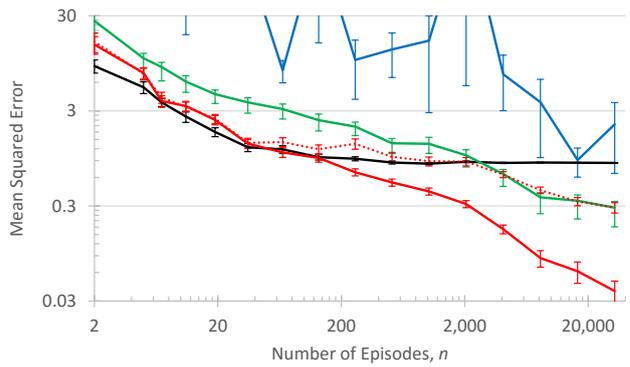

Figure 32: Hybrid, half-data.

due to an increased difficulty of determining which method to favor when they both are improving rapidly with $n$.

We also showed in Figures 31 and 32 an example where MAGIC outperformed MAGIC-B by an order of magnitude, and all previous methods (including DR) by 2–3 orders of magnitude. This exemplifies **1)** the importance of blending between importance sampling methods and purely model-based estimators using off-policy $j$-step returns, as opposed to selecting between or directly weighting WDR and AM and **2)** the power of MAGIC relative to existing estimators.

## J. Future Work

Several avenues of future work remain. Good performance of MAGIC is contingent on our ability to efficiently estimate $\Omega_n$ and $\mathbf{b}_n$, and so improved estimators for these terms could yield even better performance. For instance, if the sample mean importance weight is near zero, then the importance sampling estimators have high variance that is not captured by the sample covariance matrix that we use.

Another possible avenue of future work would be to consider how MAGIC could be applied when our fundamental assumptions are violated. For example, what should be done if the transition and reward functions of the MDP are nonstationary? Can our estimators be extended to the average reward setting? What should be done if the behavior policies are not known exactly? If the approximate model is not provided initially, but constructed from the same data that is used to produce the DR, WDR, or MAGIC estimates, will DR, WDR, and MAGIC remain strongly consistent estimators? If there are multiple approximate models available, is there a way to detect which one will work best with DR, WDR, and MAGIC?